\documentclass[sigconf]{acmart}

\newtheorem{theorem}{Theorem}

\def\BibTeX{{\rm B\kern-.05em{\sc i\kern-.025em b}\kern-.08emT\kern-.1667em\lower.7ex\hbox{E}\kern-.125emX}}

\usepackage{mathtools}
\usepackage{mathabx}
\usepackage{dsfont}
\usepackage{epstopdf}
\usepackage{xcolor}
\usepackage{tikz}
\usepackage{scalerel}
\usepackage[normalem]{ulem} %for removing underlines in the hyperref, e.g., bibliography
\usepackage{amsmath}
\usepackage[ruled,vlined]{algorithm2e}
\usepackage{graphicx}
\usepackage{subcaption}
\usepackage{caption}
\usepackage{float}
\usepackage{tabularx}
\usepackage{amsmath,amsthm}
\usepackage{bm}
\usepackage{amssymb}
\usepackage{etoolbox}
\usepackage{filecontents}
\usepackage{float}

\copyrightyear{2019} 
\acmYear{2019} 
\setcopyright{acmcopyright}
\acmConference[ICCPS '19]{10th ACM/IEEE International Conference on Cyber-Physical Systems (with CPS-IoT Week 2019)}{April 16--18, 2019}{Montreal, QC, Canada}
\acmBooktitle{10th ACM/IEEE International Conference on Cyber-Physical Systems (with CPS-IoT Week 2019) (ICCPS '19), April 16--18, 2019, Montreal, QC, Canada}
\acmPrice{15.00}
\acmDOI{10.1145/3302509.3311056}
\acmISBN{978-1-4503-6285-6/19/04}

\begin{document}
	%
	% The "title" command has an optional parameter, allowing the author to define a "short title" to be used in page headers.
	\title{Data-driven Perception of Neuron Point Process with Unknown Unknowns}
	
	%
	% The "author" command and its associated commands are used to define the authors and their affiliations.
	% Of note is the shared affiliation of the first two authors, and the "authornote" and "authornotemark" commands
	% used to denote shared contribution to the research.
	\author{Ruochen Yang}
	\authornote{Both authors contributed equally to this research.}
	\affiliation{
		\institution{Ming Hsieh Department of Electrical and Computer Engineering, University of Southern California}
		\city{Los Angeles}
		\state{California}
	}
	\email{ruocheny@usc.edu}
	
	\author{Gaurav Gupta}
	\authornotemark[1]
	\affiliation{
		\institution{Ming Hsieh Department of Electrical and Computer Engineering, University of Southern California}
		\city{Los Angeles}
		\state{California}
	}
	\email{ggaurav@usc.edu}
	
	\author{Paul Bogdan}
	\affiliation{
		\institution{Ming Hsieh Department of Electrical and Computer Engineering, University of Southern California}
		\city{Los Angeles}
		\state{California}
	}
	\email{pbogdan@usc.edu}

	%
	% By default, the full list of authors will be used in the page headers. Often, this list is too long, and will overlap
	% other information printed in the page headers. This command allows the author to define a more concise list
	% of authors' names for this purpose.

	%
	% The abstract is a short summary of the work to be presented in the article.
	\begin{abstract}
		Identification of patterns from discrete data time-series for statistical inference, threat detection, social opinion dynamics, brain activity prediction has received recent momentum. In addition to the huge data size, the associated challenges are, for example, (i) missing data to construct a closed time-varying complex network, and (ii) contribution of unknown sources which are not probed. Towards this end, the current work focuses on statistical neuron system model with multi-covariates and unknown inputs. Previous research of neuron activity analysis is mainly concerned with effects from spiking history of the target neuron and the interaction with other neurons in the system while ignoring the influence of unknown stimuli. We propose to use \textit{unknown unknowns}, which describes the effect of unknown stimuli, undetected neuron activities and all other hidden sources of error. The generalized linear model links neuron spiking behavior with past activities in the ensemble neuron system, as well as the unknown influence. We develop a maximum likelihood estimation method based on fixed-point iteration. The fixed-point iterations converge fast, and besides, the proposed methods can be efficiently parallelized to offer computational advantage especially when the input spiking trains are over long time-horizon. The developed framework provides an intuition into the meaning of having extra degrees-of-freedom in the data to support the need for unknowns. The proposed algorithm is applied to simulated spike trains and on real-world experimental data of mouse somatosensory, mouse retina and cat retina. The implementation shows a successful increase of the model likelihood with respect to the conditional intensity function, and it also reveals the convergence with iterations. Results suggest that the neural connection model with unknown unknowns can efficiently estimate the statistical properties of the process by increasing the network likelihood.
	\end{abstract}
	
	%%%%%%%%%%%
	%
	% The code below is generated by the tool at http://dl.acm.org/ccs.cfm.
	% Please copy and paste the code instead of the example below.
	%
	

	%
	% Keywords. The author(s) should pick words that accurately describe the work being
	% presented. Separate the keywords with commas.
	
	%%%%%%%%%%

	%
	% This command processes the author and affiliation and title information and builds
	% the first part of the formatted document.
	\maketitle
	\section{Introduction}
	\label{sec:intro}

	Understanding the microscopic brain activity (mechanisms) in action and in context is crucial for learning and control of the neuron behavior. Towards this end, the main purpose of studying neuron activities is to identify neuron history process and their inter-dependence in the ensemble neural system. The technique of multiple electrodes makes it possible to record and study the spiking activity of each neuron in a large ensemble system simultaneously \cite{wilson1993dynamics,brown2005theory,lewicki1998review}. At first, the study was mainly focusing on single neuron behavior and the bivariate relationship of neuron pairs or triplets, while ignoring the possible ensemble neuron effect \cite{krumin2010multivariate,brown2004multiple,okatan2005analyzing}. Later, the multivariate auto-regressive framework was introduced with more complex neuronal connections. Multiple covariates affect the spiking activity of each neuron in the system \cite{kim2011granger}. The most common covariates are the intrinsic, extrinsic and other parameters related to inputs and the environmental stimuli. Previous attempts have been made to analyze the impact of spiking history and the interaction with other neurons \cite{truccolo2005point,okatan2005analyzing,kim2011granger}. 
	
	From the mathematical perspective, the neuron activity and their spike trains are stochastic in nature and their statistics are time-varying or non-stationary. Traditionally, the neuron activity is modeled by a discrete time series event of point processes \cite{karr2017point,brown2005theory,brown2004multiple}. The likelihood method for multivariate point process is a central tool of statistical learning to model neuron spiking behaviors. The likelihood function is a variate of the parameters for the point process. These parameters are estimated from the experimental data with statistical tools \cite{brown2004multiple}. Numerical methods of linear or non-linear regression including gradient ascending, iterative reweighed least square and fixed point iteration are adopted in previous likelihood estimation study \cite{chornoboy1988maximum,okatan2005analyzing,truccolo2005point,borisyuk1985new}. However, these models are concerned with the closed system assumption and does not involve the effects of the unknown external sources.
	
	The inclusion of unknowns in the context of time-varying complex networks have shown promising results in the fractional dynamical models \cite{gauravACC2018} to represent spatio-temporal physiological signals, and making predictions for motor-imagery tasks \cite{gauravICCPS2018}. The adoption of unknown sources is also applicable to the neural spiking system. Prior work on the neural activity modeling assume that all neurons in the system can be monitored and their activities are available for mathematical modeling. The excitatory and inhibitory behavior are among the detected neurons in the system. However, in reality, the sensors and detectors can only access the neurons at the surface of monitored region. Neurons underneath the external layer and the environment stimuli, which are referred as the unknown inputs to the brain system, can also influence the neural system and contribute to the brain activity in action and in context. Related to unknown artifacts of the neuron system, a neuron model with common input, modeling the spiking trains as a point process with hidden term of a Gaussian-Markov dynamics and implementing Kalman filters has been described in \cite{kulkarni2007common}. In order to model the neural activity and reconstruct the latent dynamics of the network of neurons, \cite{macke2011empirical} describes a data-driven linear Gaussian dynamics modeling framework that accounts for the hidden inputs of the neural system. Although the above-mentioned prior work considered the importance of hidden inputs onto the neural system, they rely on specific assumptions concerning the mathematical process of the unknowns. In this paper, we propose a neural system model with more general unknown inputs, i.e., we remove restrictive dynamical assumption on the unknowns and put mild assumptions on its prior statistics. At the same time, we show that the proposed methods are computationally efficient and suitable for big size of datasets.
	
	\begin{figure*}
		\centering
		\begin{tikzpicture}[scale = 1.4]
		
		\node[anchor=south west,inner sep=0] at (0,0) {\includegraphics*[viewport=0 0 745 355, width = 6.5in, height = 3in]{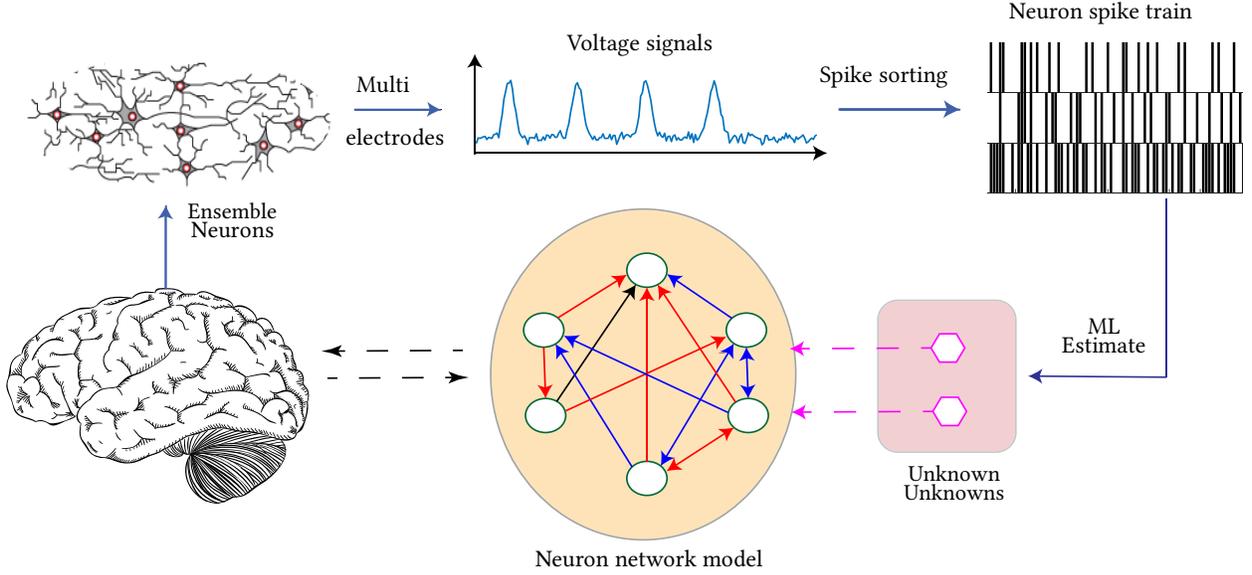}};
		
		\node[anchor=north west] at (1.7,3.3) {$\substack{\scaleto{\text{Ensemble}}{6pt}\\ \scaleto{\text{Neurons}}{6pt}}$};
		
		\node[anchor=north west] at (3.3,4.5) {Multi};
		\node[anchor=north west] at (3.2,4) {electrodes};
		
		\node[anchor=north west] at (5.3,4.9) {Voltage signals};
		
		\node[anchor=north west] at (7.7,4.6) {Spike sorting};
		
		\node[anchor=north west] at (9.5,5.2) {Neuron spike train};
		
		\node[anchor=north west] at (10,2.2) 
		{$\substack{\scaleto{\text{ML}}{6pt}\\
				\scaleto{\text{Estimate}}{6pt}}$};
		
		\node[anchor=north west] at (8.5,0.8) 
		{$\substack{\scaleto{\text{Unknown}}{6pt}\\
				\scaleto{\text{Unknowns}}{6pt}}$};
		
		\node[anchor=north west] at (5,0) {Neuron network model};
		
		\end{tikzpicture}
		
		\caption{A systematic process flow of the neuron network modeling with unknown artifacts. The neuron ensembles are probed via multiple electrodes which produce voltage signals corresponding to neuronal activities. The voltage signals are thresholded to get spike trains via spike sorting methods. The multi-neuronal spike trains are fed to maximum-likelihood estimation procedure which estimates point process based neuron network model in addition to the unknown artifacts arising due to unknown sources throughout the data collection process.}
		\label{fig:NeuronProcess}
	\end{figure*}
	
	Figure\,\ref{fig:NeuronProcess} illustrates a cyber-physical systems approach to sensing, modeling, analysis and control of neural activity \cite{lewicki1998review,brown2004multiple,buzsaki2004large,quiroga2004unsupervised}. Multiple electrodes record the neuron system activities from some area of the brain, e.g., motor cortex, somatosensory cortex. The process of spike sorting assigns spikes to specific neuron in the system. 
	Using generalized linear model framework with likelihood analysis, we analyze neuron spike trains after the spike sorting, and model the neuron system with unknown unknowns. In this paper, we propose a neuron system model with the inclusion of general unknown artifacts. The neuron spiking process has simultaneous effects from: (i) its own spiking history, (ii) the activities of other ensemble neurons, and (iii) the unknown sources. We develop computationally efficient fixed-point iteration method for the multivariate model to estimate the parameters and the unknowns which is suitable for big data size. The real-world datasets usually have enough redundancies which could be exploited to fill the gaps using unknowns. We refer to this phenomena as the extra degrees-of-freedom offered by the data. At first thought, the concept of incorporating unknowns, intuitively, seems always necessary to every dataset. However, sometimes the data does not have enough degrees-of-freedom. In such cases, a good modeling technique should be able to differentiate whether to include the unknowns or not. We have explored this phenomena in the context of real-world dataset as explained in Section\,\ref{ssec:Retina}.
	
	The paper is organized as follows. In section \ref{sec:probForm}, we introduce the neuron spiking model with the unknown artifacts assumption, and the problem definition considered in this work. Next, in Section\,\ref{sec:modelEst} we provide a maximum likelihood estimation algorithm for estimating the system model parameters. In Section\,\ref{sec:sim}, we generate artificial neuron spike trains and apply the proposed method to analyze the simulated data. In Section\,\ref{sec:realWorld}, we implement the algorithm on variety of real-world neuron spiking data and analyze the benefits. The associated challenges and interesting observations are discussed. We conclude in Section\,\ref{sec:concl} and the proofs are presented in the Appendix.
	
	\section{Problem Formulation}
	\label{sec:probForm}
	
	In this section, we first describe our point process model of neuron system in discrete time with inclusion of unknown artifacts. The monitored neuron behavior is modeled with having influence of (i) its own spiking history, (ii) other neurons activities, (iii) and the unknown unknowns. We formally describe the problem statement addressed in this work in the following subsections.
	
	\subsection{Neuron Point Process}
	\label{ssec:pointP}
	
	We consider a multi-neuron network with a total number of $C$ neurons, and assume that their spiking activities are monitored simultaneously during an observation interval $[T_{s},T_{s}+K\tau)$. The spiking interval length $\tau$ is usually small (in the orders of milliseconds), $K$ is the total number of observations and $T_{s}$ is the starting time. The neuron spiking activities are modeled as point process. Let $N_{k}^{c}$ denote the spike counting process of the $c$-th neuron, $1\leq c\leq C$, during the time interval $[T_{s},T_{s}+k\tau)$, $k=1,\hdots,K$. Also, a more useful quantity called incremental spiking count $\Delta N_k^c$ is the number of spike count fired by the neuron $c$ in time interval $[T_s+(k-1)\tau,T_s+k\tau), k=1,2,\hdots,K$, and ${\Delta}N_{1:K}^{1:C}$ is the incremental sample path of the entire monitored multi-neuron activity. 
	
	In the similar fashion, let $\Delta U_k^i$ represent the unknown artifact activity in time interval $[T_s+(k-1)\tau,T_s+k\tau)$, $k=1,2,\hdots,K$, where $i$ is the index of the unknown, $1\leq i\leq I$, and $I$ is the total number of unknowns. In what follows, we always assume that $I < C$. Moreover, we can also define $\Delta U^{1:I}_{1:K}$ as the activity path of the unknowns during the observed time horizon $[T_s,T_s+K\tau)$.
	
	The probability density function (PDF) of any $c$-th neuron with sample path $N_{1:K}^c$ can be expressed with respect to the conditional intensity function (CIF) $\lambda^c(k\vert\mathcal{H}_{k};\Theta)$, where $\mathcal{H}_{k}$ is the spiking history till $k$ time-interval and $\Theta$ are the associated parameters. The CIF $\lambda^c(k\vert\mathcal{H}_{k};\Theta)$ fully characterizes the spike trains for some neuron $c$ \cite{karr2017point, chornoboy1988maximum, brown2005theory}. The CIF can be mathematically defined as the spiking probability on time interval $[k\tau,k\tau+\Delta)$
	\begin{equation}
	\lambda^c \left( {k|\mathcal{H}_k,\Theta } \right) = \mathop {\lim }\limits_{\Delta  \to 0 } \frac{{P(N(k\tau + \Delta ) - N(k\tau))}}{\Delta},
	\label{eqn:CIF}
	\end{equation}
	\noindent where all other conditional covariates contribute to the neuron activity. In our model, the spiking history $\mathcal{H}_k=\left\{\Delta N_{1:k-1}^{1:C},\allowbreak \Delta U_{1:k-1}^{1:I}\right\}$ with $\mathcal{H}_{1} \subseteq \mathcal{H}_{2} \subseteq \hdots \subseteq \mathcal{H}_{K}$, and $\Theta$ is the parameter tuple to measure this process. 
	
	The joint conditional probability density function for the entire multi-neuron point process model can now be expressed with the CIF $\lambda^c(k|H(k);\Theta)$ \cite{brown2004multiple}. We assume that with the given spiking history $\mathcal{H}_{k}$, the activities of all neurons at $k$-th time-interval are independent, or they are conditionally independent with respect to the past. In other words, the correlation between neuronal activities appears only through the history. The joint probability of the spike train can be written as
	\begin{align}
	&P(N_{1:K}^{1:C}|\mathcal{H}_{K};\Theta) =\nonumber\\
	&\exp \left\{ \sum\limits_{c=1}^C \sum \limits_{k = 1}^K \log\lambda ^c(k\vert\mathcal{H}_{k};\Theta )\Delta N_k^c -  \tau\lambda ^c{(k\vert\mathcal{H}_{k};\Theta )}\vphantom{\sum\limits_{c=1}^C}\right\}.
	\label{eqn:jointProb}
	\end{align}
	We use generalized linear model (GLM) framework to describe the CIF along with the exponential rectification. In this model, the CIF of any neuron $c$ at time $k$ is linear function by four covariates namely: $(a)$ the spontaneous, $(b)$ the intrinsic, $(c)$ the extrinsic, and $(d)$ the unknown covariate. The CIF is formally written as 
	\begin{align}
	&\log\lambda^c(k\vert\mathcal{H}_{k};\Theta) = \alpha(c) 
	+\sum \limits_{q = 1}^{\text{min}(k-1,Q)} {\varepsilon _q}(c)\Delta N_{k-q}^c \, \nonumber\\
	&\qquad\qquad{+}\,\sum \limits_{c' = 1,c' \ne c}^C \sum \limits_{r = 1}^{\text{min}(k-1,R)} \beta _r^{c'}(c)\Delta N_{k - r}^c \nonumber\\
	&\qquad\qquad{+}\,\sum \limits_{i = 1}^I \sum \limits_{m = 1}^{\text{min}(k-1,M)} \gamma _m^i(c)\Delta U_{k - m}^i,
	\label{eqn:PPIntensity}
	\end{align}
	\noindent where $\alpha(c)$ is the spontaneous spiking rate of the neuron $c$, $\varepsilon_{q}(c)$ are the intrinsic parameters corresponding to neuron's own spiking history with a memory length of $Q$. The extrinsic parameters $\beta_{r}^{c'}(c)$ relates the effect of neurons $c', c'\neq c$ in system towards neuron $c$ with a memory length $R$. The unknown parameters $\gamma_{m}^{i}(c)$ describe the influence of unknown artifacts with the memory length of $M$. The complete parameter tuple $\Theta$ can now be formally expressed as $\Theta = (\alpha(c),\varepsilon_{q}(c), \beta_{r}^{c'}(c), \gamma_{m}^{i}(c))$ with appropriate indices of $\alpha,\varepsilon,\beta$ and $\gamma$. The CIF in equation (\ref{eqn:PPIntensity}) takes care of the interactions which are intrinsic, extrinsic as well as from unknown sources. Such interactions are often found among users' activities in the social networks, e.g., Twitter network. The modeling of such activities to decipher the user connectivity can exploit a similar CIF formulation.
	
	\subsection{Model Estimation with Unknown Unknowns}
	\label{ssec:modelEst}
	The statistical modeling of neuron spikes is restricted by the assumption that the neuron network is completely known. But even then, adding edges between neurons in the assumed network can improve the likelihood only up to some extent. At this stage, it has to be realized that closed network assumption may not be always valid and inclusion of effects of the unknown sources is necessary. The unknown source is a generic term and can include the effects of un-probed neuronal nodes, and experimental bias (e.g. instrument error, filtering process bias, physical pressure and temperature), i.e. why, they are in the most general form referred as \textit{unknown unknowns}. 
	
	To overcome the above-mentioned challenges, we propose a data-driven framework which starts from the available restricted data (i.e., few time series consisting of spiking trains of neurons) and constructs a time-varying complex network model of the neural activity that is subject to the unknown stimuli. The generic use of the term unknown unknowns is used to target all contributions that makes the data deviate from the undertaken model property.
	
	The unknown sources are assumed to be independent across time and space, and we also assume that neurons have retainment property i.e. the influence of unknown sources are not instantaneous but have some memory for each neuron. In what follows, we assume that unknown parameters $\gamma_{m}^{i}(c)$ are known for each neuron and are non-negative. Notice that this assumption of knowledge of the unknown parameters is not rigid and can be relaxed by performing a cross-validation over the set of values of parameters. Consequently, the considered problem statement in this work is formally stated as.
	
	\textbf{Problem:} \textit{Given} Spiking train data $N_{1:K}^{1:C}$ and intrinsic, extrinsic memory length $Q$ and $R$, and unknown parameters $\gamma_{m}^{i}(c)$, \textit{Estimate} the system model parameters $\Theta = (\alpha(c),\varepsilon_{q}(c), \beta_{r}^{c'}(c))$ and the unknown activities $\Delta U_{k}^{i}$.
	
	The unknown parameters are constrained to be non-negative, and we will see in Section\,\ref{sec:modelEst} that this does not restrict our modeling abilities of inhibition and excitation. The incorporation of unknowns in the model is also contingent on the assumption that the data has \textit{sufficient degrees-of-freedom to support unknowns}, in other words there are gaps that can be filled by the unknowns. In most of the real-world cases this is true, however, interestingly we have observed that in certain cases this may not happen, as illustrated in Section\,\ref{ssec:Retina}.

	\section{Statistical Spiking Model Estimation}
	\label{sec:modelEst}
	\begin{figure*}[!t]
		\setcounter{equation}{7}
		\begin{minipage}{\textwidth}
			\begin{eqnarray}
			\nu_q^{i_0\,(n+1)} &=& \nu_q^{i_0\,(n)} {\left[\frac 
				{\sum\limits_{c = 1}^{C}\sum\limits_{k = q+1}^{\text{min}(q+M,K)} \Delta N_k^c \gamma^{i_0}_{k-q}(c)+\beta}
				{\sum\limits_{c = 1}^{C}\sum\limits_{k = q+1}^{\text{min}(q+M,K)} \gamma^{i_0}_{k-q}(c)\omega^c(k\vert\Theta)\tau
					\lambda_U^c(k\vert{\nu}^{(n)})+\frac{{\nu_q^{i_0\,(n)}}}{\alpha}} \right]}^{t_q^{i_0}},\label{eqn:thm1_1}\\
			{t_q^{i_0}}&=& l\left[\frac{\sum\limits_{c = 1}^{C}\sum\limits_{k = q+1}^{\text{min}(q+M,K)} \Delta N_k^c \gamma^{i_0}_{k-q}(c)+\beta}
			{\sum\limits_{c=1}^{C}\sum\limits_{p=\text{max}(q-M+1,1)}^{\text{min}(q+M-1,K)}\sum\limits_{k=\text{max}(p+1,q+1)}^{\text{min}(p+M,q+M,K)} \gamma^{i_0}_{k-q}(c)\gamma^{i_0}_{k-p}(c)\Delta N_{k}^{c}}\right],\label{eqn:thm1_2}
			\end{eqnarray}
		\end{minipage}
		\setcounter{equation}{3}
	\end{figure*}
	%\begin{widetext}
	%	%	\setcounter{equation}{3}
	%	\begin{eqnarray}
	%	\nu_q^{i_0\,(n+1)} &=& \nu_q^{i_0\,(n)} {\left[\frac 
	%		{\sum\limits_{c = 1}^{C}\sum\limits_{k = q+1}^{\text{min}(q+M,K)} \Delta N_k^c \gamma^{i_0}_{k-q}(c)+\beta}
	%		{\sum\limits_{c = 1}^{C}\sum\limits_{k = q+1}^{\text{min}(q+M,K)} \gamma^{i_0}_{k-q}(c)\omega^c(k)\tau
	%			\lambda_U^c(k\vert{\nu}^{(n)})+\frac{{\nu_q^{i_0\,(n)}}}{\alpha}} \right]}^{t_q^{i_0}},\label{eqn:thm1_1}\\
	%	{t_q^{i_0}}&=& l\left[\frac{\sum\limits_{c = 1}^{C}\sum\limits_{k = q+1}^{\text{min}(q+M,K)} \Delta N_k^c \gamma^{i_0}_{k-q}(c)+\beta}
	%	{\sum\limits_{c=1}^{C}\sum\limits_{p=\text{max}(q-M+1,1)}^{\text{min}(q+M-1,K)}\sum\limits_{k=\text{max}(p+1,q+1)}^{\text{min}(p+M,q+M,K)} \gamma^{i_0}_{k-q}(c)\gamma^{i_0}_{k-p}(c)\Delta N_{k}^{c}}\right],\label{eqn:thm1_2}
	%	\end{eqnarray}
	%\end{widetext}
	With the spiking data and some initial knowledge of the unknown parameters $\gamma$, the goal of the estimation procedure as described in this section is to perform two tasks, first (i) estimate the system model parameters $\Theta = (\alpha(c),\varepsilon_{q}(c), \beta_{r}^{c'}(c))$, and simultaneously (ii) estimate the unknown activities $\Delta U_{1:K}^{1:I}$. To perform these two tasks, we propose an Expectation-Maximization (EM) formulation \cite{McLachlam,dempster1977maximum}. The proposed algorithm like EM is split into two parts. First, it estimates the unknown activities having some previous knowledge of the system model parameters. In the next step, the algorithm uses this estimated unknown activity values to update the system model parameters $\Theta$. These steps are repeated until convergence. The goal of the algorithm is to maximize the likelihood, and the proposed procedure being iterative will provide a maximal likelihood solution. The log likelihood associated with the current objective can be written as
	\begin{equation}
	l = \sum\limits_{c = 1}^C {\sum\limits_{k = 1}^K {\log \lambda^c(k\vert\mathcal{H}_{k};\Theta)} } \Delta N_{k}^{c} - \tau \lambda^c(k\vert\mathcal{H}_{k};\Theta).
	\label{eqn:logLikeli}
	\end{equation}
	The log likelihood in (\ref{eqn:logLikeli}) is a function of CIF, and at this point it is convenient to split the CIF in equation (\ref{eqn:PPIntensity}) into two parts as follows.
	\begin{eqnarray}
	\lambda _U^c(k\vert\Delta U) &=& \exp \left\{\sum \limits_{i = 1}^I \mathop \sum \limits_{m = 1}^{\text{min}(k-1,M)} \gamma _m^i(c)\Delta U_{k - m}^i \right\}, \label{eqn:CIF_firstPart}\\
	\omega^c(k\vert \Theta) &=& \exp \left\{ \alpha(c) + \mathop \sum \limits_{q = 1}^{\text{min}(k-1,Q)} {\varepsilon _q}(c)\Delta N_{k - q}^c \right.\nonumber\\ 
	&&\quad\left. {+}\:\mathop \sum \limits_{c' = 1,c' \ne c}^C \mathop \sum \limits_{r = 1}^{\text{min}(k-1,R)} \beta _r^{c'}(c)\Delta N_{k - r}^c \right\},
	\label{eqn:CIF_secondPart}
	\end{eqnarray}
	\noindent where $\lambda^{c}(k\vert \mathcal{H}_{k};\Theta) = \lambda_U^{c}(k \vert \Delta U)\,\omega^{c}(k\vert\Theta)$. It can be readily realized that the first part $\lambda_U^{c}(k\vert\Delta U)$ is a function of unknown activities $\Delta U$, while the second part $\omega^{c}(k\vert\Theta)$ is function of system models parameters $\Theta$. Hence, the `E' and `M' like step will alternatively update these two parts of the CIF, respectively, to maximize the log likelihood in equation (\ref{eqn:logLikeli}) iteratively. The CIF partition $\lambda _U^c(k\vert\Delta U)$ will be used for the rest of the paper in its most useful form as follows.
	\begin{equation}
	\lambda_U^{c}(k \vert \nu) = \prod\limits_{i=1}^{I}\prod\limits_{m=1}^{M}(\nu_{k-m}^{i})^{\gamma_{m}^{i}(c)},
	\label{eqn:CIF_E_part}
	\end{equation}
	\noindent where $\nu_{k}^{i} = e^{\Delta U_{k}^{i}}$. The update of unknown activities, or also $\nu_{k}^{i}$, is performed using the following result.
	\begin{theorem}
		Given neuron spikes $\Delta N^{1:C}_{1:K}$, $\omega^c(k\vert\Theta)$ from (\ref{eqn:CIF_secondPart}), time interval $\tau$, prior parameters $\alpha$, $\beta$, and the unknown parameters $\gamma$, the unknown artifacts $\Delta U_{q}^{i_{0}}$ are estimated using fixed-point iterations at each iteration index $n$ as in equation (\ref{eqn:thm1_1})-(\ref{eqn:thm1_2}), where $\nu^{i_0}_q=e^{\Delta U^{i_0}_q}$ and $l \in (0,2)$. The Maximum likelihood estimate of $\nu_{k}^{i_0}$ is denoted as $\widehat{\nu}_{k}^{i_0}$.
		\label{thm:thm_E_step}
	\end{theorem}
	
	The reason for restricting the values of unknown parameters $\gamma_{k}^{i}$ to be non-negative in Section\,\ref{ssec:modelEst} can now be realized more concretely from equation (\ref{eqn:thm1_1}) and (\ref{eqn:thm1_2}). It can be seen that the denominators of terms in both (\ref{eqn:thm1_1}) and (\ref{eqn:thm1_2}) can go negative (depending on the data) and hence the fixed-point iterations would possibly become intractable. However, as already mentioned, this does not restrict our ability to model the inhibition and excitation effects, because now it can be decided through the sign of cumulative estimated unknown activities.
	
	For the next step, we wish to update the other part of CIF written in (\ref{eqn:CIF_secondPart}) as $\omega^{c}(k\vert\Theta)$. Again, we express the $\omega^{c}(k\vert\Theta)$ in its most useful form for the rest of the paper by defining the following vectors.
	\begin{eqnarray}
	\setcounter{equation}{10}
	\mu^{c} &=&  \lbrack e^{\alpha(c)},\hdots,e^{\epsilon_n(c)},\hdots,e^{\beta_r^{c'}(c)},\hdots \rbrack,\label{eqn:muDef}\\
	Y^c(k)&=&
	\lbrack 1,\hdots,\Delta N^{c'}_{k-r},\hdots,\Delta N_{k-n}^c,\hdots\rbrack,
	\label{eqn:YDef}
	\end{eqnarray}
	\noindent where $Y^{c}(k)$ and $\mu^{c}$ are $D\times 1$ vectors, $D = 1 + Q + (C-1)R$. The $\omega^{c}(k\vert\Theta)$ can now be written as 
	\begin{equation}
	\omega^{c}(k\vert \Theta) = \prod\limits_{l=1}^{D}(\mu_{l}^c)^{Y_{l}^c(k)},
	\end{equation}
	\noindent where $\mu_{l}^c$ and $Y_{l}^c(k)$ are $l$-th component of $\mu^{c}$ and $Y^c(k)$ in (\ref{eqn:muDef}) and (\ref{eqn:YDef}) respectively.
	
	%In the M-Step, with $\Delta U^*$ derived from previous E-Step, the expectation function $Q(\theta)=E_{\Delta U\vert N;\theta }\left\{\log P(N,\Delta U;\theta)\right\}$ is approximated as the log-likelihood of $P(N\vert \Delta U^*;\theta)$. And the log-likelihood is:
	%
	%\begin{equation}
	%\begin{aligned}
	%l(\theta)=\sum\limits_{c = 1}^C {\sum\limits_{k = 1}^K{\Delta N_k^c\log{(\lambda^c(k))}-\tau\lambda^c(k)}}
	%\label{llhM}
	%\end{aligned}
	%\end{equation}
	
	\begin{theorem}[\cite{chornoboy1988maximum}]
		Given neuron spikes $\Delta N^{1:C}_{1:K}$, $\mu^c$, $Y^c(k)$ from (\ref{eqn:muDef})-(\ref{eqn:YDef}), and time interval $\tau$, the exponential of system models parameters $\mu^{c}$ as defined in (\ref{eqn:muDef}) are updated using fixed-point iterations at iteration index $n$ as follows.
		
		\begin{align}
		{\mu^c_j}^{(n+1)}&={\mu^c_j}^{(n)}\left[{\frac{\sum\limits_{k=1}^K{\Delta N_k^c Y_j^c(k)}}{\sum\limits_{k=1}^K{\tau Y_{j}^{c}(k) \left[\prod\limits_{l=1}^D{({\mu^c_j}^{(n)})^{Y^c_l(k)}}\right]}}}\right]^{\beta_j^c}, \label{eqn:thm2_1}\\
		\beta_j^c&=\frac{\sum\limits_{k=1}^K{\Delta N_k^c Y_{j}^c(k)}}{\sum\limits_{k=1}^K{\tau Y_{j}^{c}(k) \left[\prod\limits_{l=1}^D{(\widehat {\mu_l^c})^{Y_{l}^c(k)}}\right]}\left[\sum\limits_{l=1}^D{Y_{l}^{c}(k)}\right]},\label{eqn:thm2_2}
		\end{align}
		\noindent where $\widehat {\mu_l^c}$ is the maximum likelihood estimate of $\mu_l^c$.
		\label{thm:thm_M_step}
	\end{theorem}
	The denominator of $\beta_{j}^{c}$ is a variant of maximum likelihood (ML) estimate of $\mu_{l}^{c}$ which is problematic as it is not available at the time of iterations. However, an approximation with counting argument is provided in \cite{chornoboy1988maximum,chornoboy1986Thesis} which works well for the estimation problems.
	\begin{equation}
	\beta_j^c \approx\frac{\sum\limits_{k=1}^K{\Delta N_k^c Y_{j}^{c}(k)}}{\sum\limits_{k=1}^K{ Y_{j}^{c}(k) \Delta N_{k}^{c} }\left[\sum\limits_{l=1}^D{Y_{l}^{c}(k)}\right]}.
	\label{eqn:betaApprox}
	\end{equation}
	The estimation results in Theorem\,\ref{thm:thm_E_step} and Theorem\,\ref{thm:thm_M_step} are used to construct the following iterative algorithm.
	%
	%----------------------------------------------------------------------------
	%--------------------------ALG-I : EM ALGORITHM------------------------------
	%----------------------------------------------------------------------------
	%
	
	\begin{algorithm}
		{\small
			\SetKw{KwInitialize}{Initialize}
			\SetArgSty{normal}
			\KwIn{ $N_{1:K}^{1:C}$, $\gamma_{m}^{i}$ $1\leq i\leq I$, $1\leq m\leq M$ and memory lengths $Q, R$}
			\vspace*{5pt}
			\KwOut{$\Theta = (\alpha(c),\varepsilon_{q}(c), \beta_{r}^{c'}(c))$ , $1\leq c\leq C, \allowbreak 1\leq q\leq Q, \allowbreak 1\leq r\leq R$, and $\Delta U_{1:K}^{1:I}$} 
			\vspace*{5pt}
			\KwInitialize{For $t=0$, set $\Theta^{(0)}$ by using Theorem\,\ref{thm:thm_M_step} with $\Delta U_{1:K}^{1:I} \sim U[-1,1]$}
			
			\Repeat{until converge}{
				
				\textbf{`E-step'}: \\
				(i) For each $i_{0} \in [1,I]$ and $q \in [1,K]$, using $\omega^{c}(k\vert \Theta^{(j)})$ obtain $\nu_{q}^{i_0\,(j+1)}$ from Theorem\,\ref{thm:thm_E_step}  and hence $\lambda_U^{c}(k \vert \nu^{(j+1)})$;
				\\\vspace*{5pt}
				\textbf{`M-step'}: \\ 
				(i) Using $\lambda_U^{c}(k \vert \nu^{(j+1)})$ obtain $\mu^{c\,(j+1)}$ from Theorem\,\ref{thm:thm_M_step} and hence $\omega^{c}(k\vert \Theta^{(j+1)})$;
				\\
				(ii) $\Theta^{(j+1)}\leftarrow \text{log}\,\mu^{c\,(j+1)};$
				\\\vspace*{5pt}
				$j \leftarrow j + 1$\;
			}
		}
		\caption{EM algorithm}
		\label{alg:EM_alg}
	\end{algorithm}
	%--------------------------------------------------------------------------- 
	\begin{figure}[t]
		\centering
		\begin{tikzpicture}
		\node[anchor=north west,inner sep=0] at (0,0) {\includegraphics*[viewport=0 0 270 300, width = 2.2in, height = 2.3in]{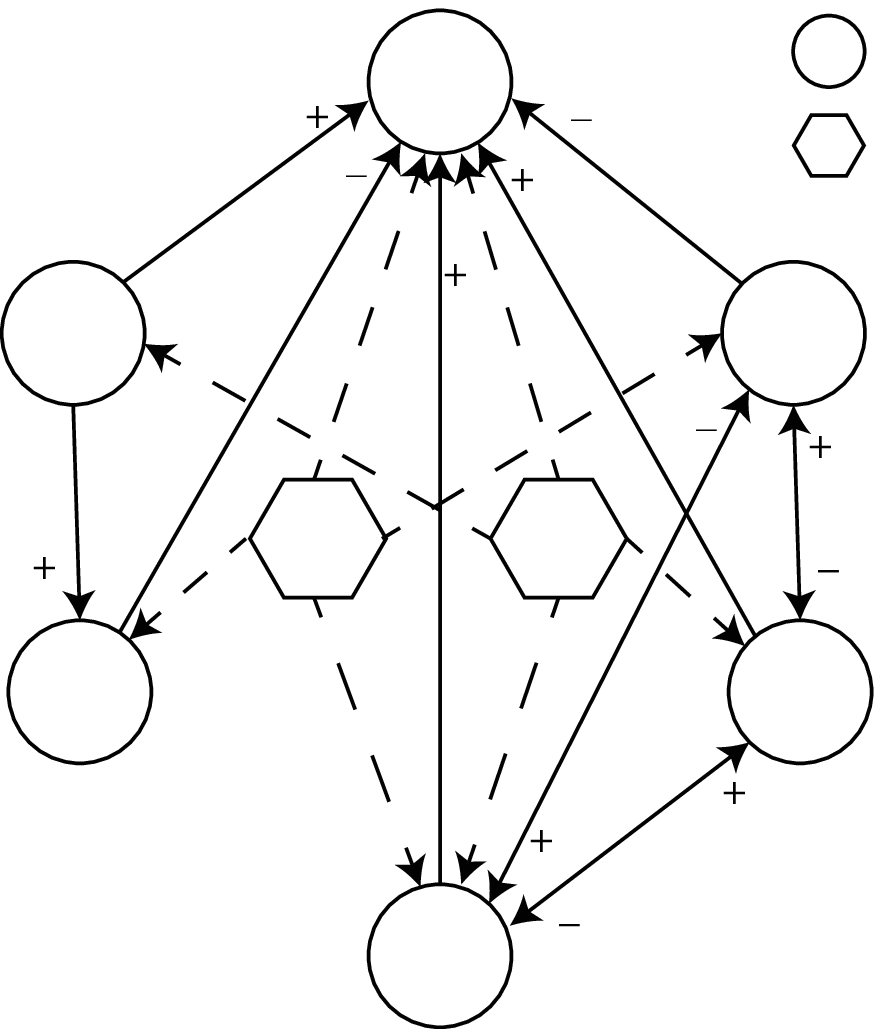}};
		\node[anchor=north west] at (2.3,-0.3) {\large{$n_{1}$}};
		\node[anchor=north west] at (0.15,-1.75) {\large{$n_{2}$}};
		\node[anchor=north west] at (0.15,-3.75) {\large{$n_{3}$}};
		\node[anchor=north west] at (2.3,-5.2) {\large{$n_{4}$}};
		\node[anchor=north west] at (4.45,-3.75) {\large{$n_{5}$}};
		\node[anchor=north west] at (4.45,-1.75) {\large{$n_{6}$}};
		
		\node[anchor=north west] at (1.6,-2.9) {\large{$u_{1}$}};
		\node[anchor=north west] at (3.05,-2.9) {\large{$u_{2}$}};
		
		\node[anchor=north west] at (5.25,-0.05) {{observed}};
		\node[anchor=north west] at (5.25,-0.60) {{unknown}};
		\end{tikzpicture}
		\caption{Neuron network assumed for the generating artificial neuronal spikes, with six observed neurons and two unknown artifacts. The excitation and inhibition effects are indicated by $+$ and $-$ respectively for each directed edge.}
		\label{fig:neuronModelSim}
	\end{figure}
	\begin{figure*}[!ht]
		\centering
		\begin{subfigure}[b]{0.25\linewidth}
			\includegraphics*[viewport=0 0 485 360, width = 1.75in, height = 1.6in]{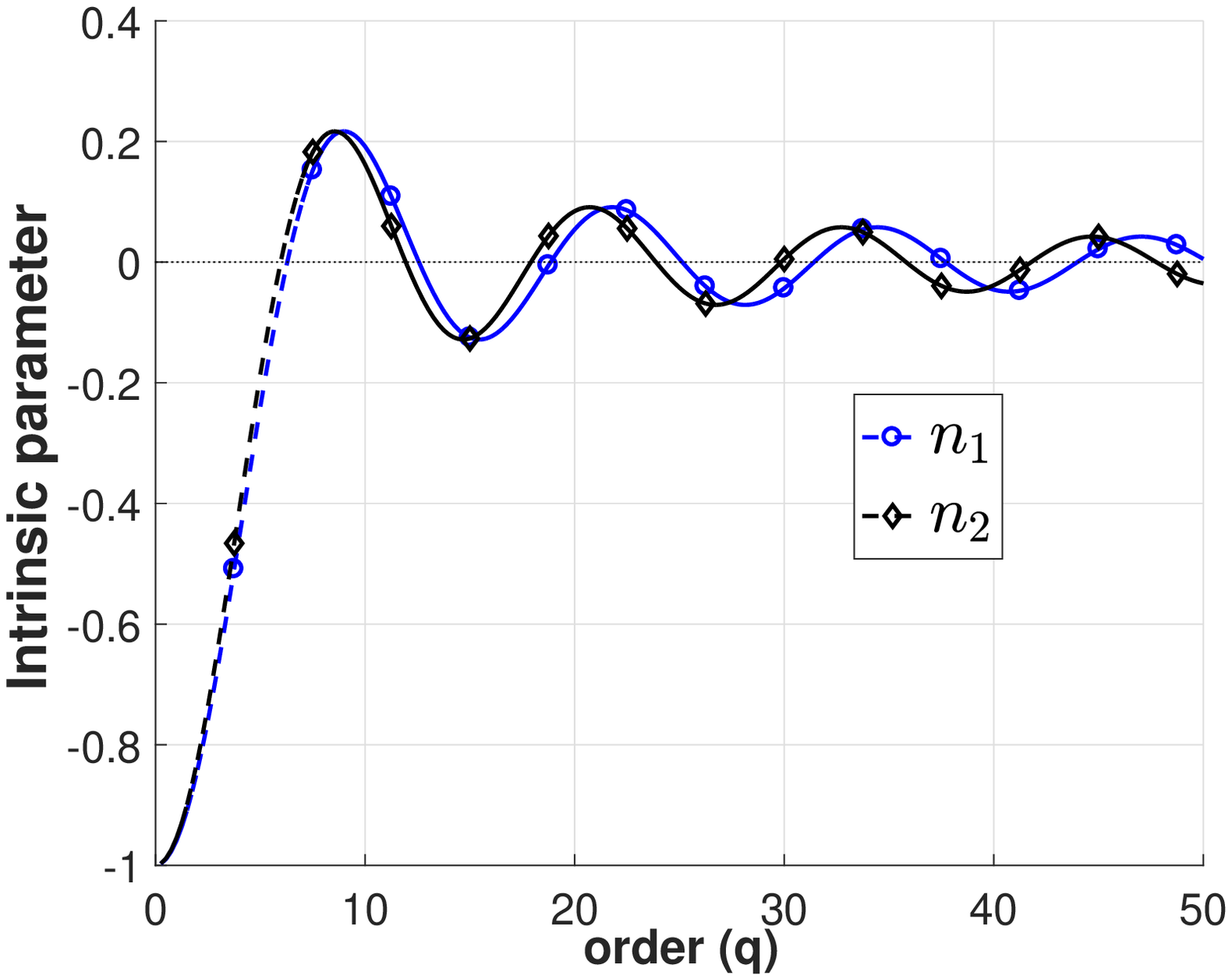}
			\caption{}
			\label{sfig:intr12}
		\end{subfigure}\hfill
		\begin{subfigure}[b]{0.25\linewidth}
			\includegraphics*[viewport=0 0 485 360, width = 1.75in, height = 1.6in]{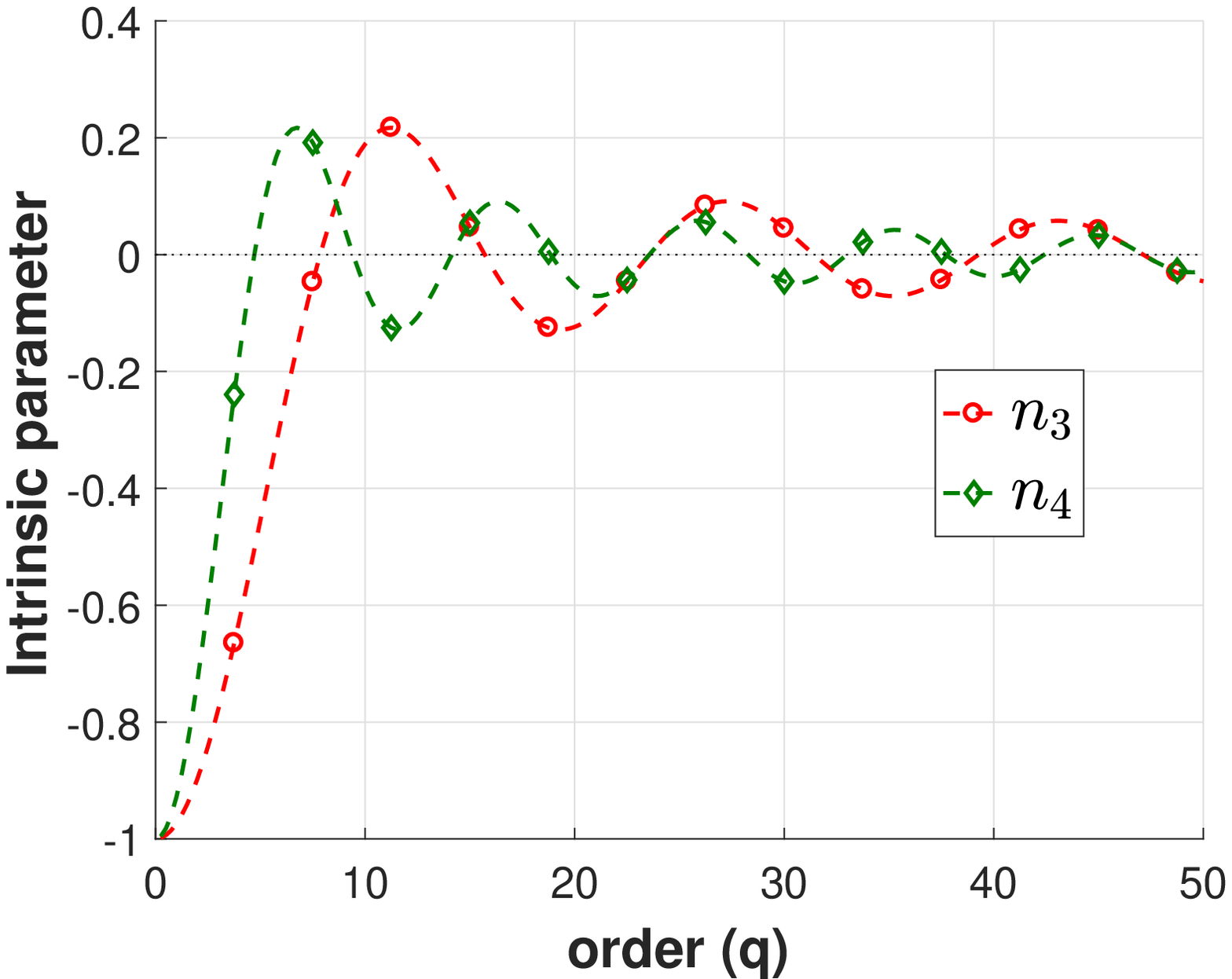}
			\caption{}
			\label{sfig:intr34}
		\end{subfigure}\hfill
		\begin{subfigure}[b]{0.25\linewidth}
			\includegraphics*[viewport=0 0 485 360, width = 1.75in, height = 1.6in]{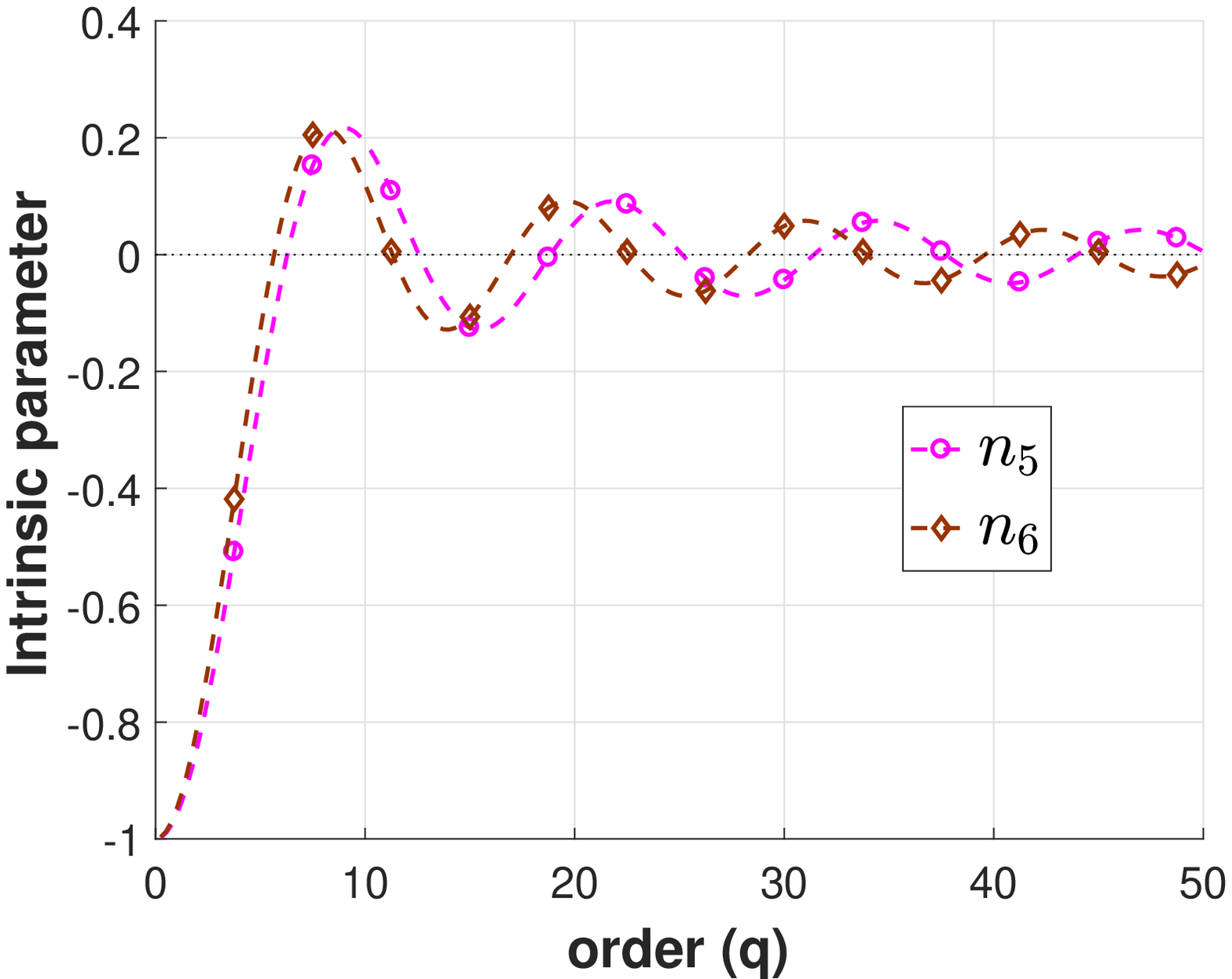}
			\caption{}
			\label{sfig:intr56}
		\end{subfigure}\hfill
		\begin{subfigure}[b]{0.25\linewidth}
			\includegraphics*[viewport=0 0 485 360, width = 1.75in, height = 1.6in]{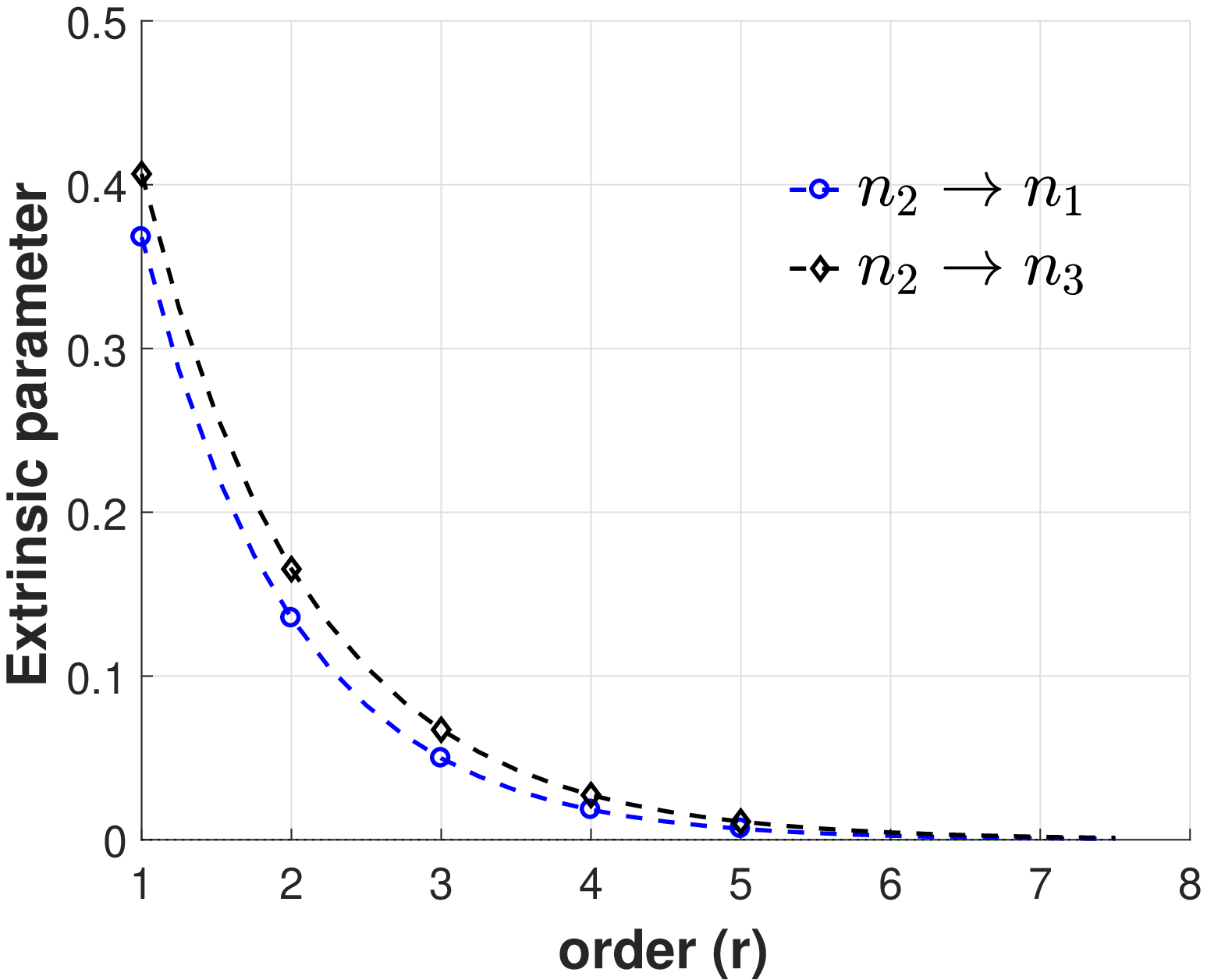}
			\caption{}
			\label{sfig:extr2}
		\end{subfigure}
		\begin{subfigure}[b]{0.25\linewidth}
			\includegraphics*[viewport=0 0 485 360, width = 1.75in, height = 1.6in]{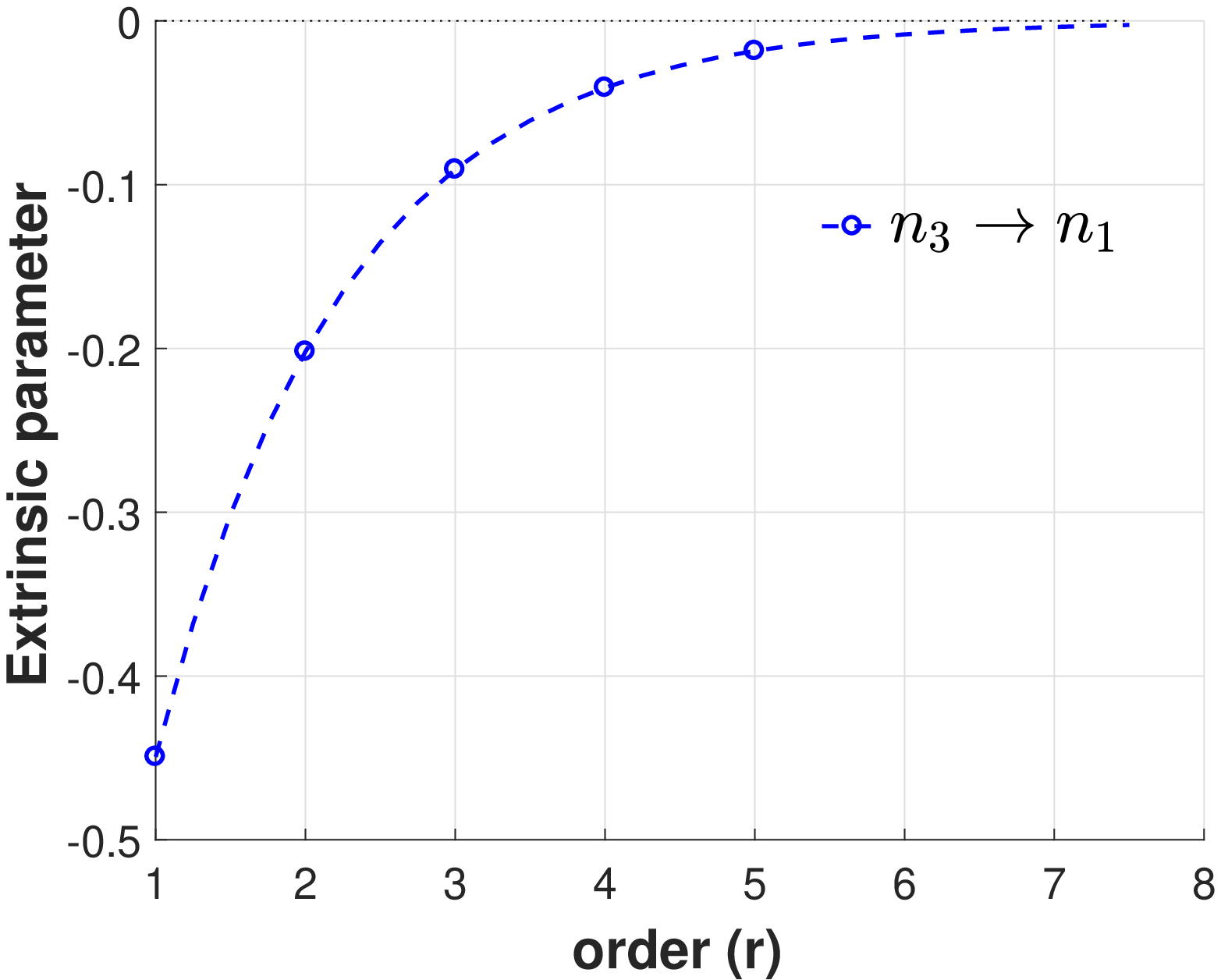}
			\caption{}
			\label{sfig:extr3}
		\end{subfigure}\hfill
		\begin{subfigure}[b]{0.25\linewidth}
			\includegraphics*[viewport=0 0 485 360, width = 1.75in, height = 1.6in]{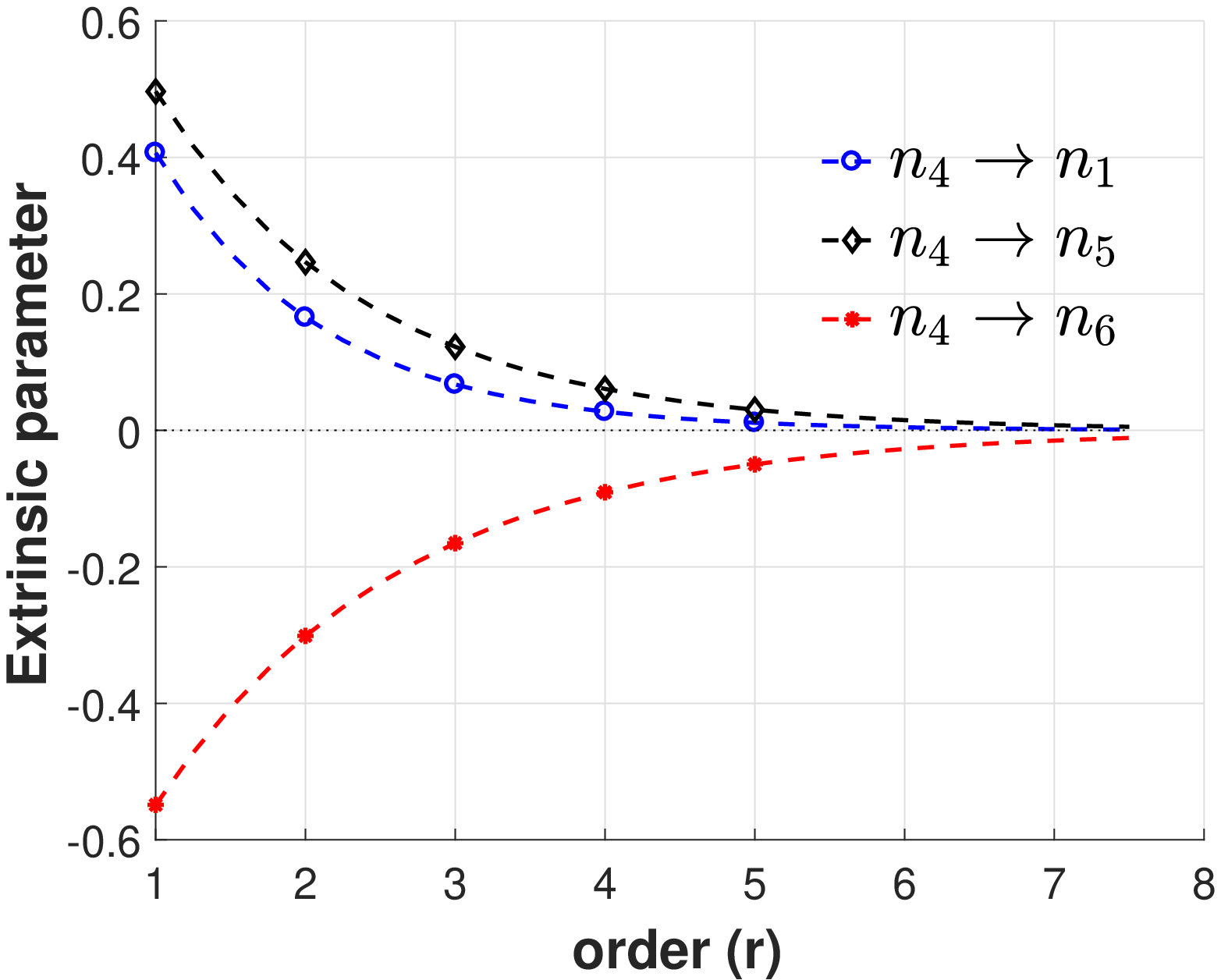}
			\caption{}
			\label{sfig:extr4}
		\end{subfigure}\hfill
		\begin{subfigure}[b]{0.25\linewidth}
			\includegraphics*[viewport=0 0 485 360, width = 1.75in, height = 1.6in]{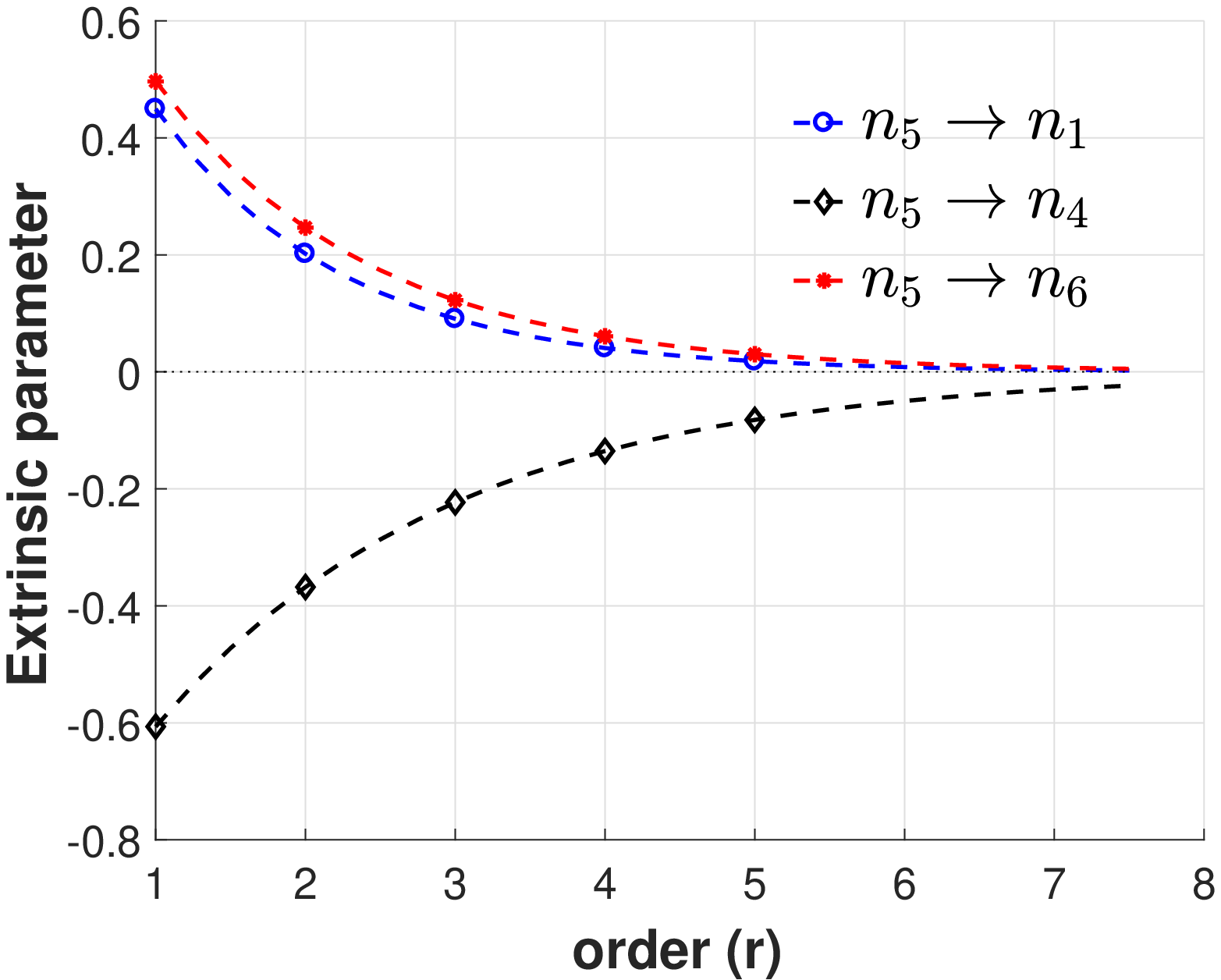}
			\caption{}
			\label{sfig:extr5}
		\end{subfigure}\hfill
		\begin{subfigure}[b]{0.25\linewidth}
			\includegraphics*[viewport=0 0 485 360, width = 1.75in, height = 1.6in]{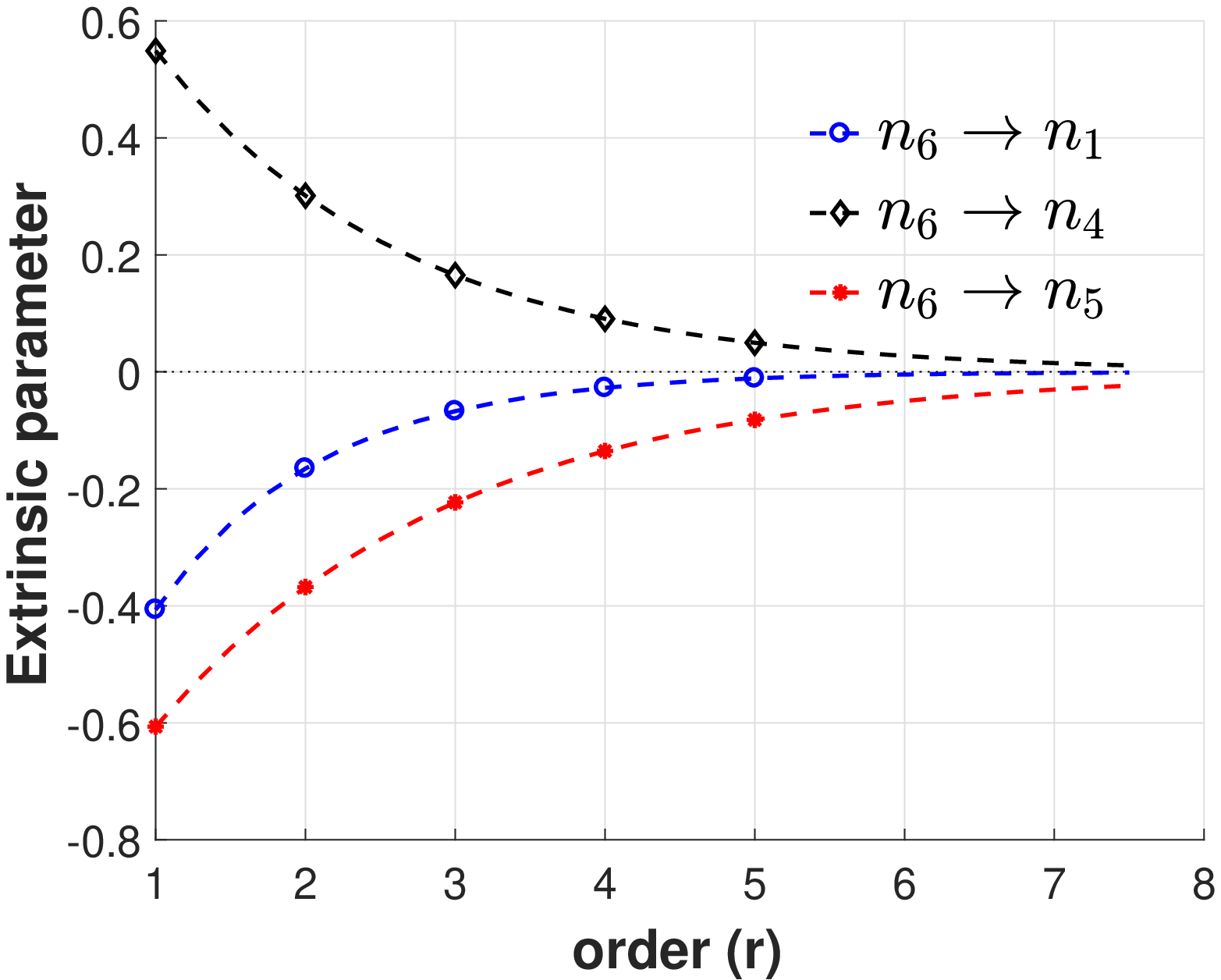}
			\caption{}
			\label{sfig:extr6}
		\end{subfigure}
		\caption{Neuron point process model parameters for the considered artificial network. The intrinsic parameters $\varepsilon_{q}(c)$ for $1\leq c\leq 6$ are shown in (a)-(c). The extrinsic parameters $\beta_{r}^{c'}(c)$ for each directed edge $n_{c'}\rightarrow n_{c}$ in Figure\,\ref{fig:neuronModelSim}  are shown in (d)-(h). The excitation and inhibition is taken care of through signs of corresponding extrinsic parameters.}
		\label{fig:simPara}
	\end{figure*}
	It should be noted that in each fixed-point iteration, both in Theorem\,\ref{thm:thm_E_step} and Theorem\,\ref{thm:thm_M_step}, the value of exponent $t_{q}^{i_0}$ and $\beta_{j}^{c}$ is constant with respect to the iteration index $n$. Similarly, the numerators of the fixed-point functions are constant. Hence, they need to be computed only once per EM update and lot of computations can be saved. It is also important to note that another big advantage offered by proposed fixed-point iterations is that they are independent across time index $q$ and unknown index $i_{0}$, therefore they can be implemented in parallel using current \textit{multi-threading/multi-processing} architectures. This make the computations very fast especially when we have large size of the data. On the other hand, the existing Kalman smoothing techniques \cite{kulkarni2007common} have dependence across time and has to be computed serially. The fixed-point iterations and hence EM iterations are in the closed-form, and they are computationally efficient. The convergence is fast, as we will see in Section\,\ref{sec:sim} and Section\,\ref{sec:realWorld}. The choice of scalar $l$ in equation (\ref{eqn:thm1_2}) can play an important role in convergence rate and hence can be taken as an input parameter as well.
	
	The proposed techniques developed in this section are tested on simulated as well as real-world datasets in Section\,\ref{sec:sim} and Section\,\ref{sec:realWorld}, respectively.
	
	\section{Simulation Experiment}
	\label{sec:sim}
	
	We now demonstrate the applicability of the proposed neuron spiking model estimation technique with unknown unknowns. The estimation process, as detailed in the Algorithm\,\ref{alg:EM_alg}, is applied to an artificially generated spiking data which is explained in the following parts.
	
	\begin{figure}
		\centering
		\begin{subfigure}[b]{0.5\linewidth}
			\includegraphics*[viewport=0 0 485 367, width = 1.6in, height = 1.4in]{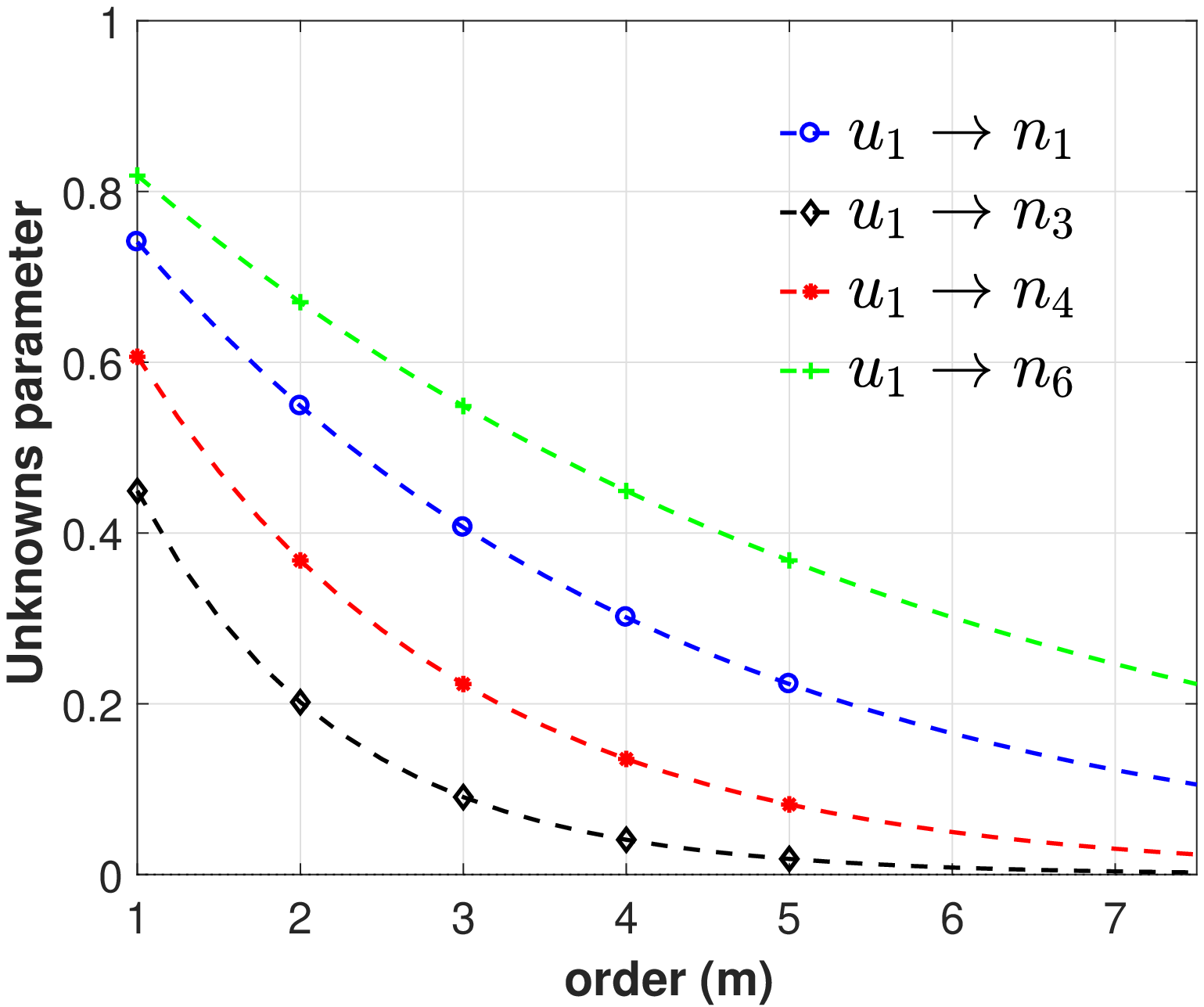}
			\caption{}
			\label{sfig:uPara1}
		\end{subfigure}\hfill
		\begin{subfigure}[b]{0.5\linewidth}
			\includegraphics*[viewport=0 0 485 367, width = 1.6in, height = 1.4in]{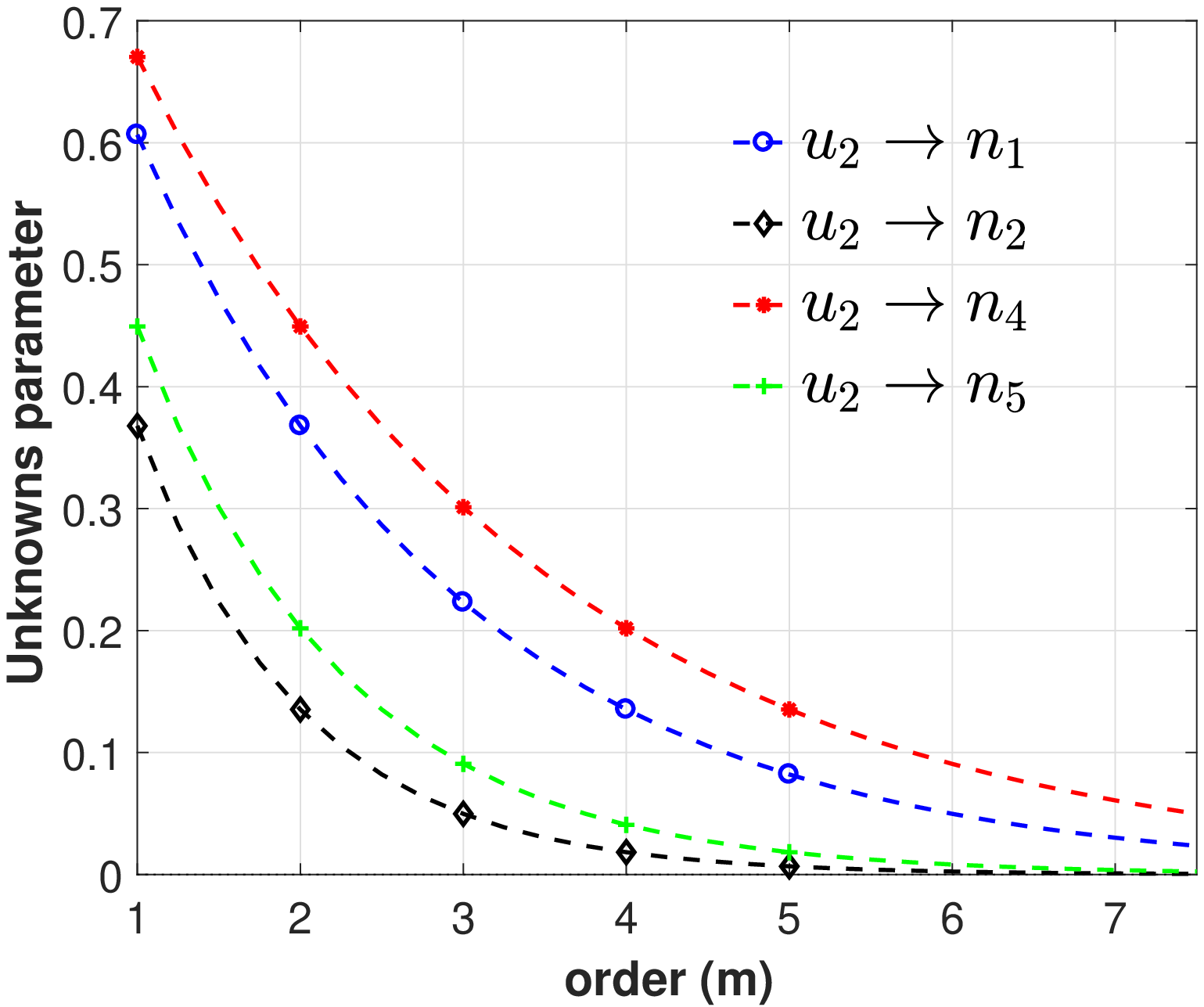}
			\caption{}
			\label{sfig:uPara2}
		\end{subfigure}
		\caption{The unknown node parameters $\gamma_{m}^{i}(c)$ for two unknown sources according to the considered directed edges in Figure\,\ref{fig:neuronModelSim} with contribution of $u_{1}$ in (a) and $u_{2}$ in (b).}
		\label{fig:simUPara}
	\end{figure}
	
	\subsection{Artificial Neuron Network}
	\label{ssec:artNN}
	
	An artificial neuron network in Figure\,\ref{fig:neuronModelSim} is designed with a total of six neurons. Each neuron is assumed to be influenced by $(a)$ its intrinsic activity, $(b)$ extrinsic effects via other neurons in the network, and $(c)$ unknown artifacts. The contribution of unknown sources is quantized by having two unknown nodes $u_{1}$ and $u_{2}$. The extrinsic effect is modeled by having directed edges among neurons as shown in Figure\,\ref{fig:neuronModelSim}. For example, neuron $n_{2}$ excites $n_{3}$, and the excitation effect is indicated using $+$, also, neuron $n_{6}$ inhibits $n_{1}$, and the inhibition effect is marked through $-$ sign. The contribution of unknown nodes is indicated by the corresponding directed edges from $u_{i}$ to $n_{j}$ in the network.
	
	The parameters required for defining the intensity function of the spiking point-process as in equation (\ref{eqn:PPIntensity}) are selected as follows: The values of spontaneous firing rate $\alpha$ for six neurons are selected as, $\alpha=[3.5,2.4,2.0,3.0,2.8,2.5]$. The intrinsic effect memory length is selected as $Q=50$ for all neurons. The previous spiking history of each neuron can have excitatory as well as inhibitory effects on the future spiking activity. Also, with the increasing length of the history, the effects get mitigated. The initial values of the intrinsic parameters are negative to model the refractory period of the neurons after firing a spike \cite{truccolo2005point,berry1998refractoriness}. To collectively model these effects, we describe the intrinsic parameters using sinc function. The intrinsic parameter values are selected as shown in Figure\,\ref{sfig:intr12}-\ref{sfig:intr56}. Next, the extrinsic effect as indicated via directed edges in Figure\,\ref{fig:neuronModelSim} is quantized though parameters $\beta_{r}^{c'}(c)$ for each $n_{c'}\rightarrow n_{c}$. The extrinsic memory length is fixed as $R = 5$ for each directed edge. We have used exponential decay functions to model the parameters and the values are shown in Figure\,\ref{sfig:extr2}-\ref{sfig:extr6}. Next, the unknown sources contribution memory length is fixed as $M = 5$ for both sources $u_{1}$ and $u_{2}$. The parameters associated with unknown excitations are always taken to be positive, as explained in Section\,\ref{ssec:modelEst}, and the undertaken values are shown in Figure\,\ref{fig:simUPara}. Finally, with these parameter values we can now proceed to the spike generation process.
	\begin{figure}
		\centering
		\includegraphics*[viewport=50 0 660 505, width = 3.3in, height = 2.5in]{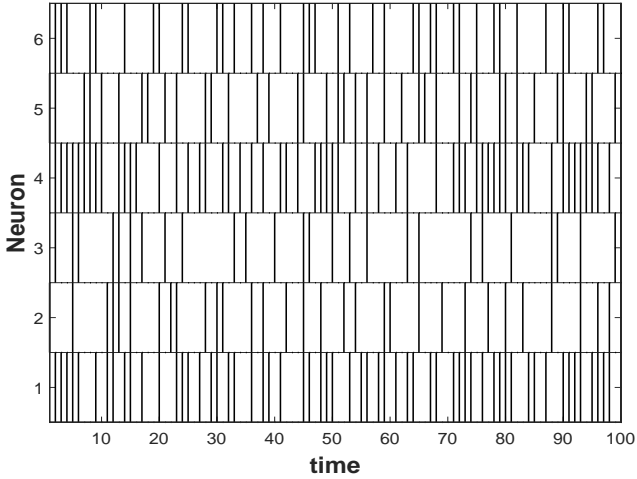}
		\caption{Simulated neuron spike trains for six neurons with system model parameters as described in Section\,\ref{ssec:artNN}.}
		\label{fig:spikeSim}
	\end{figure}
	\subsection{Spike Generation with Unknown contributions}
	\label{ssec:spikeGenSim}
	
	The multi-neuron spike generation can be performed by recursively compute the conditional intensity parameter, or firing rate, and adding the contributions of the unknown sources. As mentioned in Appendix, we have used log-Gamma distribution for $\Delta U_{k}^{i}$, and independent and identically distributed (iid) samples of size $[I\times K]$ are generated using the PDF in equation (\ref{eqn:logGammaDistr}). The parameter values are taken to be $\alpha = 1$ and $\beta = 50$, and the samples are then mean centered to zero. The spike train generation procedure is similar to \cite{kim2011granger} and for the sake of completeness, we briefly write the steps as follows:
	\begin{itemize}
		\item At step $k$, compute the conditional intensity function from equation (\ref{eqn:PPIntensity}) using the system model parameters as stated in Section\,\ref{ssec:artNN}, and unknown sources contribution $\Delta U_{k}^{i}$ from (\ref{eqn:logGammaDistr}).
		\item Generate a uniform random number $u\sim U[0,1]$ and if $u < \tau\lambda^c(k|\mathcal{H}_k,\theta)$, then there is spike for $c$-th neuron at time interval $k$
		\item  Repeat with recursively computing conditional intensity function from equation (\ref{eqn:PPIntensity}), until desired length of spike train.
	\end{itemize}
	\noindent where the value of $\tau$ is taken to be $0.05$, and for the $k$-th step, the time-interval in consideration is $[T_{s}+(k-1)\tau,T_{s}+k\tau)$. For simulations, we have generated a total of $K = 500$ multi-neuron spikes. The spike train obtained using above procedure for all six neurons in the first $100$ time-intervals is as shown in Figure\,\ref{fig:spikeSim}. We now apply the proposed technique of model estimation with unknown unknowns in the next section.
	
	\subsection{Model estimation}
	\label{ssec:modelEstSim}
	
\begin{figure}
		\centering
		\includegraphics*[viewport=60 0 640 490, width = 2.7in, height = 2.1in]{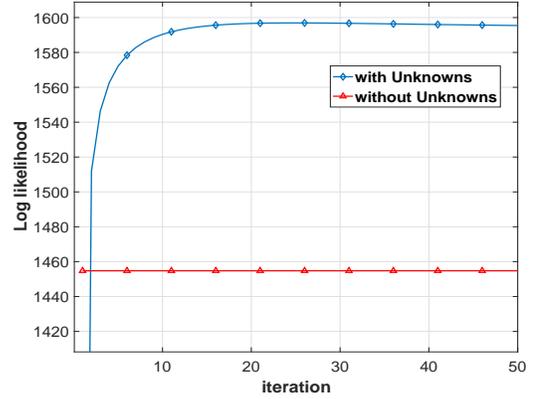}
		\caption{Log likelihood at each iteration using Algorithm\,\ref{alg:EM_alg} on the simulated spike trains.}
		\label{fig:LLH_Sim}
	\end{figure}

	The artificially generated spike train in Figure\,\ref{fig:spikeSim} with the parameters for unknown sources as shown in Figure\,\ref{fig:simUPara}  is used as an input for the Algorithm\,\ref{alg:EM_alg}. The algorithm recursively computes the unknown contribution $\Delta U_{k}^{i}$ and then update the system model parameters. The iterations are fixed-point based and the convergence is fast. The log likelihood plot (with constants ignored) is shown in Figure\,\ref{fig:LLH_Sim}. The `without unknowns' case is computed with setting unknowns to zero and, for consistency, we run the `without unknowns' case with maximum `M-step' iterations occurring in the proposed `with unknowns' approach. As expected, there is sharp increase in the likelihood in the first few iterations, due to incorporation of unknowns, and it is observed that few EM iterations are sufficient for convergence. The proposed approach results in better log likelihood as compared to the case of no unknowns.
	
	The proposed approach shows good results on the simulated data, but there are other challenges that need to be addressed especially when the data is coming from real-world experiments. We will address some of them by considering a variety of real-world datasets in the Section\,\ref{sec:realWorld}.
	
	\section{Real-World Data: Case Study}
	\label{sec:realWorld}
	In this section, we explore various neuron spiking data recorded through real-world experiments. Each dataset poses some challenges as compared to the simulated dataset considered in the Section\,\ref{sec:sim}. We describe the datasets and discuss the results in the following sections.
	
	\subsection{Mouse Somatosensory Spiking Data}
	\label{ssec:SSC}
	
\begin{figure}
		\centering
		\includegraphics*[viewport=50 0 660 505, width = 3.3in, height = 2.5in]{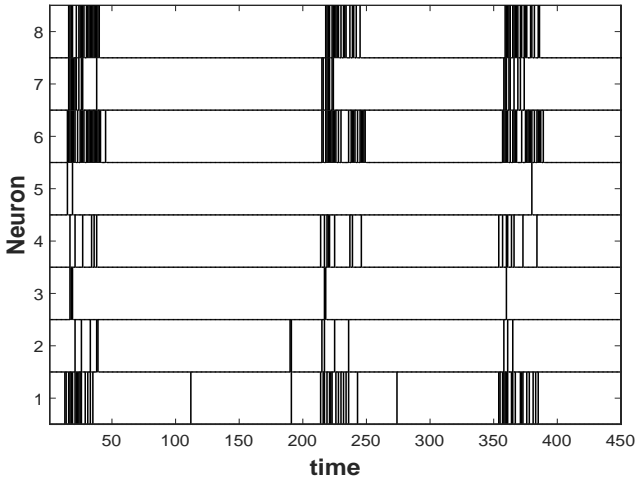}
		\caption{Multi-neuron spike train for Somatosensory data with a total of eight neurons.}
		\label{fig:spikeSSC3}
	\end{figure}

	The mouse somatosensory spiking dataset is recorded using a 512-channel multi-electrode array system in an organotypic cultures of dorsal hippocampus and somatosensory cortex sections \cite{SSC_1,SSC_2, SSC_3, SSC_4, SSC_5,SSC_6}. The cultures were not stimulated and the dataset represents spontaneous activity. The spikes were sorted using Principle Component Analysis (PCA). The dataset is downloaded from \cite{SSC_Data}, where a total of $25$ datasets are available, and hundreds of neurons were probed to have the spiking data. In this work, we have taken a total of $8$ neurons to study the inter-neuronal interactions with unknown unknowns. The spiking activity of the considered neurons is as shown in Figure\,\ref{fig:spikeSSC3}. 
	
\begin{figure}
		\centering
		\includegraphics*[viewport=50 0 640 490, width = 2.7in, height = 2.3in]{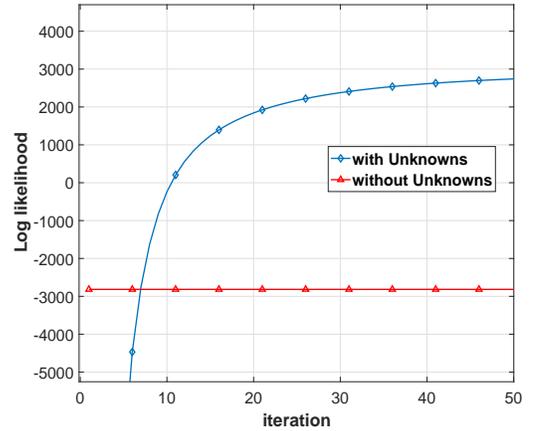}
		\caption{Log likelihood at each iteration using Algorithm\,\ref{alg:EM_alg} on the Somatosensory spike trains dataset.}
		\label{fig:LLH_SSC3}
	\end{figure}

	The inter-spiking interval is having a large value for spikes at some times, and it is very small for others. Therefore, while modeling such datasets it is possible that the assumed model for CIF in (\ref{eqn:PPIntensity}) may not apply in its original form. The proposed technique like any other data-driven estimators can work only if there are samples corresponding to the associated parameters. For example, when the data is too sparse in some time-intervals then the denominator term in equation (\ref{eqn:betaApprox})
	\begin{equation*}
	\sum\limits_{k=1}^K{ Y_{j}^{c}(k) \Delta N_{k}^{c} }\left[\sum\limits_{l=1}^D{Y_{l}^{c}(k)}\right],
	\end{equation*}
	will go to zero. It is an indication that there is no data corresponding to the $j$\,th term in (\ref{eqn:muDef}) and hence it cannot be recovered. A simple modification would be to increase the spiking interval length $\tau$ and take $\Delta N_{k}^{c}$ as accumulated counts in that enlarged interval. The proposed Algorithm\,\ref{alg:EM_alg} is then applied to the spike train and the log likelihood (with constants ignored) is shown in Figure\,\ref{fig:LLH_SSC3}. We observe that the convergence is fast and the likelihood increase sharply in the beginning. We also observe that the resulting log likelihood is better as compared to the case of not having unknowns in the model.
	
	\subsection{Mouse Retina Spiking Data}
	\label{ssec:Retina}
\begin{figure}
		\centering
		\includegraphics*[viewport=50 0 665 505, width = 3.3in, height = 2.5in]{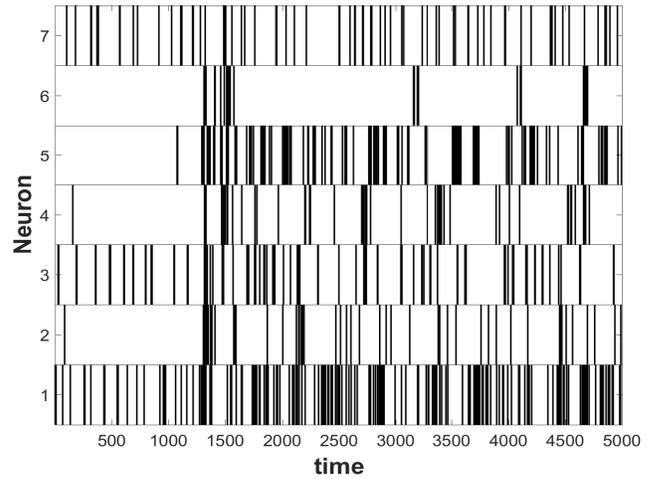}
		\caption{Multi-neuron spike train for Mice Retina data with a total of seven neurons.}
		\label{fig:spike_ret_d1}
	\end{figure}

	The mouse retina dataset contains neuronal responses of retinal ganglion
	cells to various visual stimuli recorded in the isolated retina from lab mice (Mus Musculus) using a 61-electrode array. The visual stimuli were displayed on a gamma-corrected cathode-ray tube monitor and projected on the photoreceptor layer of the retina (magnification $8.3 \mu m$/pixel; luminance range 0.5-3.8 mW/m$^2$) from above through a custom-made lens system \cite{Retina_Lefebvre4141}. Extracellular action potentials were recorded (sampling rate $10$ kHz) and single
	units were identified by a semi-automated spike-sorting algorithm. The dataset is downloaded from \cite{Retina_Data} which comprises $16$ retinal preparations. The spike trains for a total of seven neurons is shown in Figure\,\ref{fig:spike_ret_d1}.
	
\begin{figure}
		\centering
		\includegraphics*[viewport=65 0 640 510, width = 2.7in, height = 2.3in]{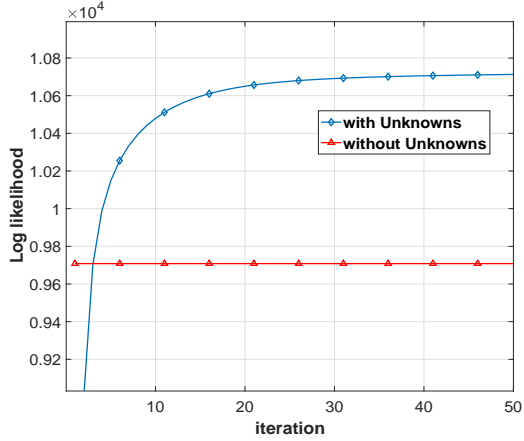}
		\caption{Log likelihood at each iteration using Algorithm\,\ref{alg:EM_alg} on the Mice Retina spike trains dataset.}
		\label{fig:LLH_ret_d1}
	\end{figure}

	The concept of having unknown sources is better in the sense that it provides flexibility to fill the gaps in the real-world data and the assumed model by exploiting the extra degrees-of-freedom that the data might have. However, we have observed in this dataset that this may not be true always, and sometimes unknown contributions are not necessary. In particular to our model, this can be realized when the denominator of equation (\ref{eqn:thm1_2})
	\begin{equation*}
	\sum\limits_{c=1}^{C}\sum\limits_{p=\text{max}(q-M+1,1)}^{\text{min}(q+M-1,K)}\sum\limits_{k=\text{max}(p+1,q+1)}^{\text{min}(p+M,q+M,K)} \gamma^{i_0}_{k-q}(c)\gamma^{i_0}_{k-p}(c)\Delta N_{c}^{k},
	\end{equation*}
	\noindent goes to zero. This indicates that there is not enough degrees-of-freedom in the data (and given values of unknown parameters $\gamma_{k}^{i}(c)$) to estimate the $\Delta U_{q}^{i_0}$. In such cases, we can set the unknown activities to zero for the corresponding $q, i_{0}$ indices, or in other words assume that there is no requirement of unknowns at $q$-th time step for $i_{0}$-th unknown source. The log likelihood (with constants ignored) after this adjustment on applying Algorithm\,\ref{alg:EM_alg} is shown in Figure\,\ref{fig:LLH_ret_d1}. We observe that the likelihood increases quickly and convergence is fast. The resulting log likelihood is also better than the model estimation procedure without having unknowns.
	
	\subsection{Cat Retina Spiking Data}
	\label{ssec:catRetina}
	
\begin{figure}
		\centering
		\includegraphics*[viewport=50 0 665 505, width = 3.3in, height = 2.5in]{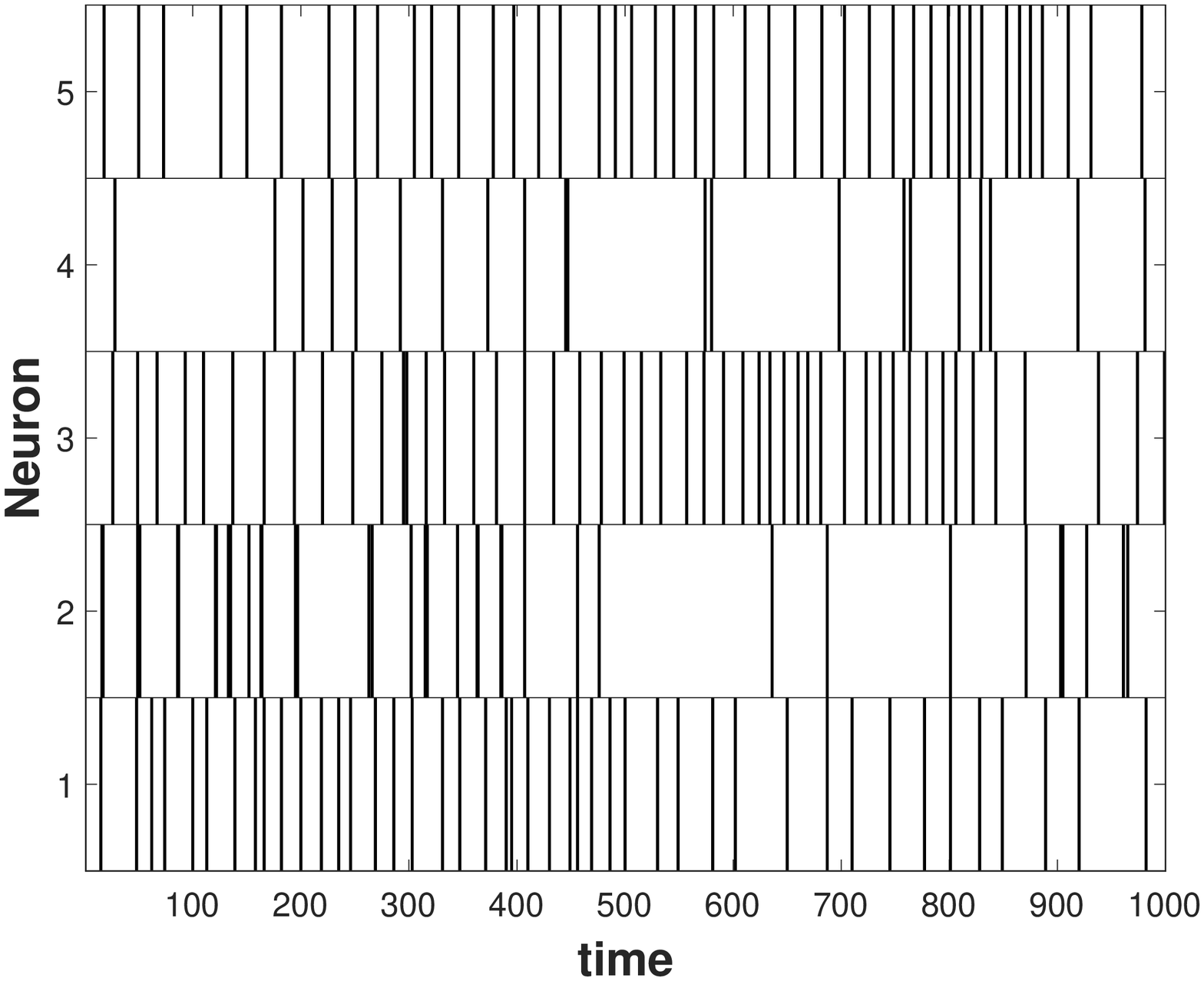}
		\caption{Multi-neuron spike train for Cat Retina data with a total of five neurons.}
		\label{fig:spike_cat_Retina}
	\end{figure}

	The cat retina dataset records spontaneous activities of the mesencephalic reticular formation (MRF) neurons of head-restrained cat to investigate
	their dynamic properties during sleep and waking \cite{Cat_Retina_1,Cat_Retina_2}. The behavioral states were classified into one of the three following states which continued for a relatively long period (several hundred seconds): $(i)$ slowwave sleep (SWS); $(ii)$ paradoxical sleep (PS); and $(iii)$ highly attentive states (BW) with sustained attention to a pair of small birds in a cage. We have taken the spike train data of PS in this work. The spike train for a total of five probed neurons is shown in Figure\,\ref{fig:spike_cat_Retina}.
	
\begin{figure}
		\centering
		\includegraphics*[viewport=75 10 660 510, width = 2.7in, height = 2.3in]{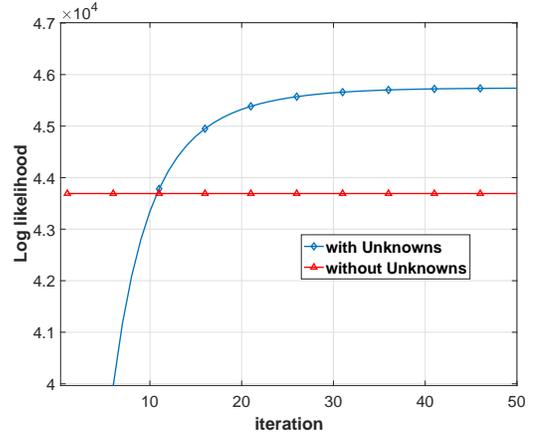}
		\caption{Log likelihood at each iteration using Algorithm\,\ref{alg:EM_alg} on the Cat Retina spike trains dataset.}
		\label{fig:LLH_cat_Retina}
	\end{figure}

	The cat retina dataset also has long and short inter-spiking intervals, and hence similar to the analysis of mouse somatosensory dataset in Section\,\ref{ssec:SSC}, we enlarge the spiking window to care of this effect. The proposed methods are applied to estimate the model parameters and the estimation procedure of Algorithm\,\ref{alg:EM_alg} converges fast. The output log likelihood (with constants ignored) is shown in Figure\,\ref{fig:LLH_cat_Retina} where we observe that the initial jump of likelihood is sharp and it gets saturated very soon. In addition, we also observe that the resulting likelihood is better than the case with having no unknowns in the model.
	
	\section{Conclusion}
	\label{sec:concl}
	
	We have presented a multi-variate neuron system framework with the inclusion of unknown inputs. The single neuron activity is affected by its own spiking history, the influence from other neurons, and the unknown stimuli. The statistical framework is built on the non-stationary likelihood of the multi-channel point process, which is characterized by the conditional intensity function with a generalized linear combination of the covariates. The proposed method aims at increasing the likelihood of the data, and the inclusion of unknown unknowns to minimize the discrepancies between the data and the model. The developed algorithm is based on fixed-point iterations and converges quickly. The proposed fixed-point method offers advantages of independence across time and hence can be computed in parallel using modern multi-core processors. Experiments on simulated spike trains and on the real-world datasets of mouse somatosensory, retinal and cat retina show promising results in terms of likelihood maximization. We have observed interesting insights into degrees-of-freedom offered by the data which sometimes suggest not to use unknown unknowns.
	
	The proposed mathematical framework is general for non -stationary discrete time-series in the form of spike trains with missing data (or gaps) and the contribution of unknown sources. The developed techniques are computationally efficient especially when the data is recorded over long time-horizon. While we relied on a log-gamma distribution as the prior for unknown artifacts, we plan to investigate unknown unknowns phenomena that exhibit non-Gaussian statistical or multifractal behavior \cite{YuankunICCPS2016}. The future research will also focus on exploring applications in the domain of pattern identification in social networks of opinion dynamics, smart cities monitoring for chemical and threat (explosive) detection and identification/localization, etc. A critical challenge with such datasets is the huge data size, and the computational advantages offered by the proposed techniques can make the statistical inference tractable.
	
	%The proposed mathematical framework is general for non -stationary discrete time-series in the form of spike trains with missing data (or gaps) and the contribution of unknown sources. The developed techniques are computationally efficient especially when the data is recorded over long time-horizon. While we have chosen log-gamma distribution as the prior of unknown artifacts, we will also explore other possible models in further study. The future research will also focus on exploring applications in the domain of pattern identification in social networks of opinion dynamics, smart cities monitoring for chemical and threat (explosive) detection and identification/localization, etc. A critical challenge with such datasets is the huge data size, and the computational advantages offered by the proposed techniques can make the statistical inference tractable.
	
	\section*{Acknowledgment}
	
	We thank reviewers for their thorough comments and suggestions for the paper. The authors are thankful to Prof. Satish Iyengar, University of Pittsburgh, and Prof. Mitsuaki Yamamoto, Tohoku Social Welfare University, for providing the Cat Retina neuron spiking dataset. The authors gratefully acknowledge the support by the U.S. Army Defense Advanced Research Projects Agency (DARPA) under grant no. W911NF-17-1-0076, the DARPA Young Faculty Award under grant no. N66001-17-1-4044 support, and the National Science Foundation Career award under grant no. CPS/CNS-1453860. The views, opinions, and/or findings contained in this article are those of the authors and should not be interpreted as representing the official views or policies, either expressed or implied by the Defense Advanced Research Projects Agency, the Department of Defense or the National Science Foundation.
	
	\nocite{truccolo2005point}
	\nocite{gauravACC2018,gauravICCPS2018}
	%\bibliographystyle{IEEEtran}
	%\bibliography{IEEEabrv,neuronPP}
	
	\bibliographystyle{ACM-Reference-Format}
	\bibliography{neuronPP}
	
	%\clearpage
	%\balancecolsandclearpage
	%\newpage
	
	% 
	% If your work has an appendix, this is the place to put it.
	\appendix
	
	\section{Proof of Theorem\texorpdfstring{\,\ref{thm:thm_E_step}}{}}
	\begin{proof}
		
		The Expectation step of the algorithm is concerned with computation of $P(\Delta U_{k}^{i}\vert N_{1:K}^{1:C};\theta)$ which is not computationally\,\,\,\, tractable in various cases. In this work, we will approximate it as $P(\Delta U_{k}^{i}\vert N_{1:K}^{1:C};\theta) = 1_{\Delta U_{k}^{i} = \Delta \widehat{U}_k^i}$ (this is sometimes referred as Hard EM \cite{Murphy}), where
		\begin{equation}
		\Delta \widehat{U}_k^i = \mathop {\arg \max }\limits_{\Delta U_k^i} \log P(\Delta U_k^i\vert N_{1:K}^{1:C};\theta).
		\label{eqn:dUargmax}
		\end{equation}
		
		The equation (\ref{eqn:dUargmax}) can be expanded using Bayesian formulation, and the choice of the prior distribution of $\Delta U_k^i$ is critical. Intuitively, $\Delta U_k^i$ represents all the hidden sources, including undetected neuron activity and environmental stimuli. Unlike binary spike count $\Delta N^c_k$, the unknown artifacts $\Delta U^i_k$ don't necessarily have to be discrete and non-negative, and are real values without further constraint. While there can be wide selections of priors, in this work we are motivated by using conjugate priors approach \cite{raiffa1974applied} to have computable expressions. Therefore, we assume the log-Gamma prior for $\Delta U_k^i$ as it collaborates well with the existing CIF model. The log-Gamma distribution \cite{demirhan2011multivariate} can be mathematically written as
		\begin{equation}
		P(\Delta U_{1:K}^{1:I};\theta) = \prod \limits_{i=1}^{I} \prod \limits_{k=1}^{K}
		\frac{e^{\beta \Delta U_k^i}e^{-e^{\Delta U_k^i/\alpha}}}
		{\alpha ^ \beta \Gamma (\beta)}, \Delta U_k^i \in \mathbb{R},
		\label{eqn:logGammaDistr}
		\end{equation}
		\noindent where $\alpha$ is the shape parameter and $\beta$ is the scale parameter. We also assume that the unknown artifacts behavior are independent and identical distributed across time and for each unknown sources. Therefore, the equation (\ref{eqn:dUargmax}) can be expanded as 
		\begin{align}
		\Delta \widehat{U}_{1:K}^{1:I} &= \mathop {\arg \max }\limits_{\Delta U_{1:K}^{1:I}} \log P(\Delta U_{1:K}^{1:I}|N_{1:K}^{1:C};\theta )\nonumber\\
		&= \mathop {\arg \max }\limits_{\Delta U_{1:K}^{1:I}} \left\{\log P(N_{1:K}^{1:C}|\Delta U_{1:K}^{1:I};\theta ) + \log P(\Delta U_{1:K}^{1:I};\theta )\right\}\nonumber\\
		&\stackrel{(a)}{=} \mathop {\arg \max }\limits_{\Delta U_{1:K}^{1:I}} \sum_{c=1}^{C}\sum_{k=1}^{K}\left\{{\Delta N_k^c \log \lambda_U^c(k)-\omega^c(k) \tau \lambda_U^c(k)}\right\} \nonumber\\
		&\qquad {+}\,\sum_{i=1}^{I} \sum_{k=1}^{K} \left\{\beta\Delta U^i_k-\frac{1}{\alpha}e^{\Delta U^i_k}\right\},
		\label{eqn:dU-argmaxExpand}
		\end{align}
		\noindent where in (a) we have used the definition of $\lambda_U^c(k)$  and $\omega^c(k)$ from equation (\ref{eqn:CIF_firstPart}) and (\ref{eqn:CIF_secondPart}), respectively. Now for the maximization, the equation (\ref{eqn:dU-argmaxExpand}) is an unconstrained optimization problem and we solve it by setting the partial derivatives with respect to $\nu_q^{i_0}=e^{\Delta U_q^{i_0}}$ to be zero for each $q=1,2,...,K$,  and $i_0=1,2,...,I$. Therefore, we obtain
		\begin{equation}
		\begin{aligned}
		&\beta-\frac{\nu_q^{i_0}}{\alpha}+\sum_{c=1}^{C}\sum_{k=q+1}^{\text{min}(q+M,K)}\Delta N_k^c\gamma_{k-q}^{i_0}(c)\\
		&\qquad{-}\,\sum_{c=1}^{C}\sum_{k=q+1}^{\text{min}(q+M,K)}\omega^c(k)\tau\gamma_{k-q}^{i_0}(c)\lambda_U^c(k\vert\nu)=0.
		\label{eqn:Ederivatives}
		\end{aligned}
		\end{equation}
		The maximum likelihood (ML) estimate of $\nu_q^{i_0}$ is computed with a fixed-point iterative method. We rearrange the terms in equation (\ref{eqn:Ederivatives}) with additional exponent $t_q^{i_0}$ and build the fixed-point function as following.
		\begin{align}
		&G^{i_0}_q(\nu) = \nu_q^{i_0} \times \nonumber\\
		&{\left[\frac 
			{\sum\limits_{c = 1}^{C}\sum\limits_{k = \text{max}(q+1,1)}^{\text{min}(q+M,K)} \Delta N_k^c \gamma^{i_0}_{k-q}(c)+\beta}
			{\sum\limits_{c = 1}^{C}\sum\limits_{k = \text{max}(q+1,1)}^{\text{min}(q+M,K)} \gamma^{i_0}_{k-q}(c)\omega^c(k)\tau
				\lambda_U^c(k\vert{\nu})+\frac{{\nu_q^{i_0}}}{\alpha}} \right]}^{t_q^{i_0}},
		\label{eqn:GDef}
		\end{align}
		\noindent where the fixed-point iterations would be $\nu_q^{i_0\,(n+1)} = G^{i_0}_q(\nu^{(n)})$ at iteration $n$. The exponent $t_q^{i_0}$ need to be chosen carefully such that the fixed-point iterations would converge to the ML solution. We follow the procedure as mentioned in \cite{peters1976numerical} to prove that $G^{i_0}_q(\nu)$ is a local contraction and hence find the value of $t_q^{i_0}$ such that the fixed-point iterations are convergent.
		
		Let us denote the ML estimate $\widehat\nu_{q}^{i_0}$ as the solution of fixed point equation $\nu_q^{i_0}=G_q^{i_0}(\nu)$. Now $G^{i}(\cdot)$ is a local contraction if $\left\lVert{\nabla G^{i}(\widehat \nu)}\right\rVert$ $<1$, where ${\nabla G^{i}(\widehat \nu)}$ is a matrix such that 
		\begin{equation}
		{\left[\nabla G^{i}(\widehat \nu)\right]}_{q,p}=\frac{\partial G_q^i(\nu)}{\partial \nu_p^i}|_{\nu = \widehat \nu}.
		\label{eqn:gradG}
		\end{equation}
		The norm $\left\lVert{\nabla G^{i}(\widehat \nu)}\right\rVert$ is less than $1$ if and only if the spectral radius $\rho(\nabla G^{i}(\widehat \nu))<1$ \cite{pattern1973classification}. After writing the $\nabla G^{i}(\widehat \nu)$ in (\ref{eqn:gradG}) using (\ref{eqn:GDef}) and upon setting 
		\begin{equation}
		t_q^{i_0} = l\frac{n_{q}^{i_0}}{d_{q}^{i_0}},
		\end{equation}
		\noindent where, $0< l < 2$ and
		\begin{align}
		n_{q}^{i_0} &= \sum\limits_{c = 1}^{C}\sum\limits_{k = q+1}^{\text{min}(q+M,K)} \Delta N_k^c \gamma^{i_0}_{k-q}(c)+\beta,\\
		d_{q}^{i_0} &= \sum\limits_{c=1}^{C}\sum\limits_{p=\text{max}(q-M+1,1)}^{\text{min}(q+M-1,K)}\sum\limits_{k=\text{max}(p+1,q+1)}^{\text{min}(p+M,q+M,K)} \gamma^{i_0}_{k-q}(c)\nonumber\\
		&\qquad\times\gamma^{i_0}_{k-p}(c)\omega^c(k)\tau
		\lambda_U^c(k\vert\widehat{\nu}),
		\end{align}
		\noindent we obtain
		\begin{equation}
		\nabla G^{i}({\widehat \nu})=I-B^i,
		\end{equation}
		\noindent where the matrix $B^i$ is non-negative and it has a positive eigenvector $\widehat{\nu}^i$ with positive eigenvalue $l$. Now, using the  Perron-Frobenius theorem \cite{householder1964theory}, given the condition that $l$ is the positive eigenvalue of $B^{i}$ with positive eigenvector, the absolute value of all other eigenvalues are less or equal than $l$. Thus, the spectral radius $\rho(\nabla G^{i}({
			\widehat \nu}))=|1-l|<1$. Hence, the fixed-point equations are convergent. It should be noted that the denominator expression $d_{q}^{i_0}$ is depending on the ML estimate $\widehat{\nu}$ which is not available during iterations. We approximate the $d_{q}^{i_0}$ using the similar counting arguments of \cite{chornoboy1988maximum} to write the final expression as in equation (\ref{eqn:thm1_2}).
	\end{proof}
\end{document}